\documentclass[default,iicol]{sn-jnl}
\usepackage[algo2e,ruled,linesnumbered]{algorithm2e}
\usepackage{array}
\usepackage{hhline}
\hypersetup{
    citecolor=cvprblue,linkcolor=red
}
\usepackage{booktabs}
\usepackage{colortbl}
\usepackage[accsupp]{axessibility}
\usepackage[capitalise]{cleveref}
\crefname{section}{Sec.}{Secs.}
\Crefname{section}{Section}{Sections}
\crefname{table}{Tab.}{Tabs.}
\Crefname{table}{Table}{Tables}
\crefname{figure}{Fig.}{Figs.}
\Crefname{figure}{Figure}{Figures}
\definecolor{cvprblue}{rgb}{0.21,0.49,0.74}
\definecolor{MyGray}{rgb}{0.85, 0.85, 0.85}
\newcommand{\tb}[1]{\textbf{#1}}
\newcommand{\ul}[1]{\underline{#1}}
\newcommand{\rb}[1]{\rotatebox{90}{#1}}
\newcommand{\ti}[1]{\textit{#1}}

\jyear{2022}%

\theoremstyle{thmstyleone}%
\theoremstyle{thmstyletwo}%

\theoremstyle{thmstylethree}%

\begin{document}

\title[Article Title]{Hard-normal Example-aware Template Mutual Matching for Industrial Anomaly Detection}


\author[1]{\fnm{Zixuan} \sur{Chen}}\email{chenzx3@mail2.sysu.edu.cn}

\author*[1,2,3]{\fnm{Xiaohua} \sur{Xie}}\email{xiexiaoh6@mail.sysu.edu.cn}

\author[1,2,3]{\fnm{Lingxiao} \sur{Yang}}\email{yanglx9@mail.sysu.edu.cn}

\author[1,2,3]{\fnm{Jian-Huang} \sur{Lai}}\email{stsljh@mail.sysu.edu.cn}

\affil[1]{
    \orgdiv{School of Computer Science and Engineering}, \orgname{Sun Yat-sen University}, \city{Guangzhou}, \postcode{510006}, \state{Guangdong}, \country{China}
}
\affil[2]{
    \orgdiv{Guangdong Province Key Laboratory of Information Security Technology}, \country{China}
}
\affil[3]{
    \orgdiv{Key Laboratory of Machine Intelligence and Advanced Computing}, \orgname{Ministry of Education}, \country{China}
}




\abstract{Anomaly detectors are widely used in industrial manufacturing to detect and localize unknown defects in query images.
These detectors are trained on anomaly-free samples and have successfully distinguished anomalies from most normal samples.
However, hard-normal examples are scattered and far apart from most normal samples, and thus they are often mistaken for anomalies by existing methods.
To address this issue, we propose \tb{H}ard-normal \tb{E}xample-aware \tb{T}emplate \tb{M}utual \tb{M}atching (HETMM), an efficient framework to build a robust prototype-based decision boundary.
Specifically, \ti{HETMM} employs the proposed \tb{A}ffine-invariant \tb{T}emplate \tb{M}utual \tb{M}atching (ATMM) to mitigate the affection brought by the affine transformations and easy-normal examples.
By mutually matching the pixel-level prototypes within the patch-level search spaces between query and template set, \ti{ATMM} can accurately distinguish between hard-normal examples and anomalies, achieving low false-positive and missed-detection rates.
In addition, we also propose \ti{PTS} to compress the original template set for speed-up.
\ti{PTS} selects cluster centres and hard-normal examples to preserve the original decision boundary, allowing this tiny set to achieve comparable performance to the original one.
Extensive experiments demonstrate that \ti{HETMM} outperforms state-of-the-art methods, while using a 60-sheet tiny set can achieve competitive performance and real-time inference speed (around 26.1 FPS) on a Quadro 8000 RTX GPU.
\ti{HETMM} is training-free and can be hot-updated by directly inserting novel samples into the template set, which can promptly address some incremental learning issues in industrial manufacturing.}




\maketitle

\begin{figure*}[!t]
    \centering
    \includegraphics[width=\textwidth]{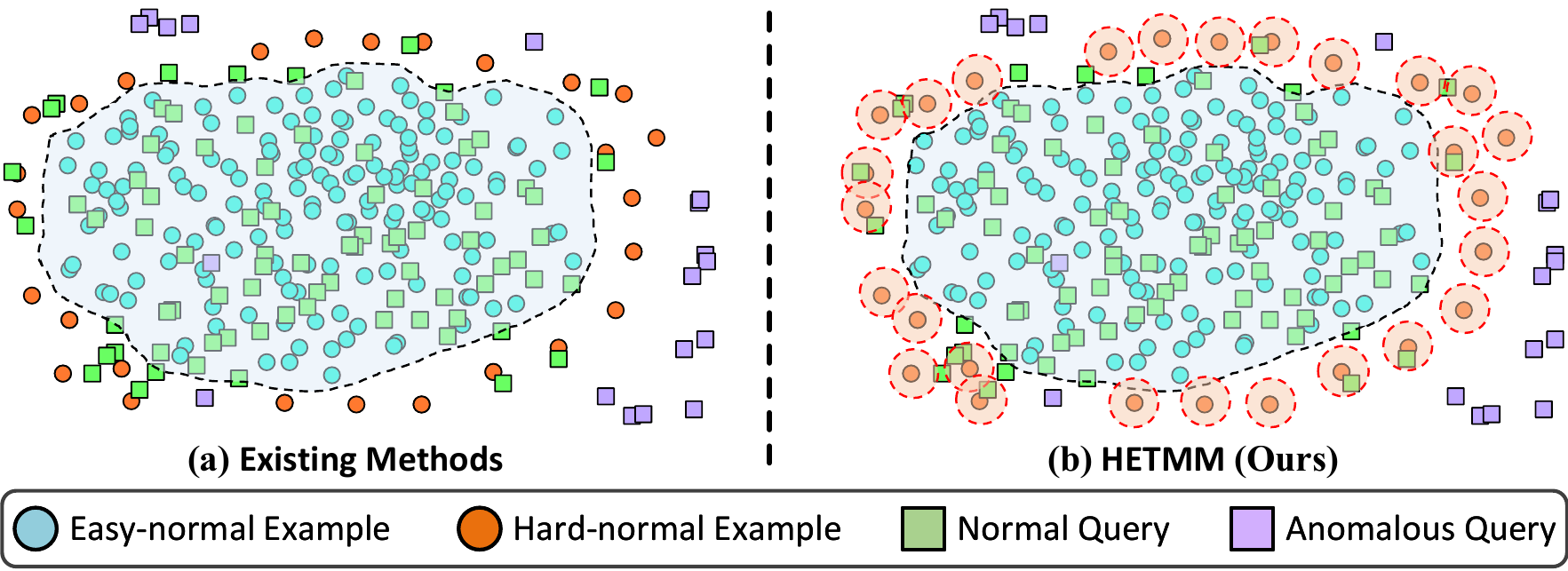}
    \caption{
 Visualization of training data (ball) and queries (cube) via t-SNE \cite{tsne}.
 Visually, existing methods' decision boundaries are dominated by the overwhelming number of easy-normal examples (blue balls).
 Hence, the normal queries (green cubes) near the hard-normal examples (orange balls) are prone to be erroneously identified as anomalies (purple cubes), resulting in a high false-positive or missed-detection rate.
 To address this issue, we propose \ti{HETMM} to construct a robust prototype-based decision boundary, which can accurately distinguish hard-normal examples from anomalies.
 }
    \label{fig:motivation}
\end{figure*}

\section{Introduction}
Industrial anomaly detection aims to detect and localize unknown anomalies in industrial images.
It is a long-standing challenge that has received widespread attention in the field of industrial vision.
Since anomalies are rare, collecting sufficient anomaly images to train a multi-class classification model is difficult.
Therefore, industrial anomaly detection is a one-class classification problem using the collected normal samples.
Earlier methods assumed that the distribution of anomalies is significantly different from normal samples.
Tax \emph{et al.} \cite{SVDD} constructed a decision boundary using hand-crafted features extracted from normal samples to discover the anomalous ones.
However, the distinction between anomalous and normal samples at the image level is often subtle.
To address this, reconstruction-based methods, such as \cite{AnoGAN,SSIM-AE,US,MKD,MVTECLOCO}, have been trained to differentiate each query pixel by pixel-level reconstruction error to localize anomalous regions accurately.
Recently, template-based methods such as \cite{SPADE, Padim, patchcore} have achieved superior performance in industrial anomaly detection.
These methods first construct the template set by aggregating normal features extracted by a pre-trained model and then identify anomalies by template matching, \emph{i.e.,} calculating the difference between queries and the template set.
Pixel-level template matching methods such as \cite{SPADE} require the query object to be aligned with the template set, while others employ template matching at a higher level, such as \cite{patchcore}, which operates at the patch level.
However, the above methods are prone to mistake some normal samples for anomalies, leading to many false alarms.

\cref{fig:motivation} illustrates that the normal samples can be divided into two distinct categories: easy- and hard-normal examples.
The normal samples that are clustered into several regions are deemed as easy-normal examples. 
In contrast, the normal samples scattered and far apart from the easy-normal examples are categorized as hard-normal examples.
Because the number of easy-normal examples far exceeds that of the hard-normal ones, this imbalanced distribution makes it difficult for most methods to distinguish between hard-normal examples and anomalies, resulting in high false-positive and missed-detection rates.

In this paper, we propose a simple yet efficient framework named \tb{H}ard-normal \tb{E}xample-aware \tb{T}emplate \tb{M}utual \tb{M}atching (HETMM) to address the above-mentioned issues.
Specifically, \ti{HETMM} aims to build a robust prototype-based boundary to distinguish between hard-normal examples and anomalies, which consists of two novel modules: \tb{A}ffine-invariant \tb{T}emplate \tb{M}utual \tb{M}atching (ATMM) and \tb{P}ixel-level \tb{T}emplate \tb{S}election (PTS).
\ti{ATMM} mutually explores the pixel-level prototypes within the patch-level search spaces between query and template set, which can accurately distinguish between hard-normal examples and anomalies.
Thus, \ti{ATMM} achieves good robustness against the affections brought by the affine transformations and easy-normal examples, resulting in much lower false-positive and missed-detection rates.
It is also worth noting that \ti{ATMM} can detect most types of anomalies, including structural and logical anomalies \cite{MVTECLOCO}, which is a significant advantage compared to most anomaly detectors that struggle to detect logical anomalies.
Additionally, to meet the speed-accuracy demands in practical production, we propose \ti{PTS} to accelerate the module inference.
Like most template-based methods, the computational cost of \ti{ATMM} is linear to the template size.
Therefore, \ti{PTS} selects some significant prototypes from the original template set to form a tiny one.
Unlike mainstream approaches that only collect cluster centres, \ti{PTS} further selects hard-normal examples to maintain the original decision boundaries.

The main contributions are summarized as follows:

\begin{itemize}
    \item We propose \tb{H}ard-normal \tb{E}xample-aware \tb{T}emplate \tb{M}utual \tb{M}atching (HETMM), a simple yet efficient framework that can construct a robust prototype-based decision boundary to distinguish hard-normal examples from anomalies.
  
    \item We propose \tb{A}ffine-invariant \tb{T}emplate \tb{M}utual \tb{M}atching (ATMM) module to alleviate the affection brought by the affine transformations and easy-normal examples.
 We also propose \tb{P}ixel-level \tb{T}emplate \tb{S}election (PTS) to streamline the original template set by selecting cluster centres and hard-normal examples to form a tiny set, enabling the models to meet various speed-accuracy demands in industrial production.
 Notably, \ti{PTS} further selects hard-normal examples as prototypes to maintain the original decision boundary, where these hard-normal prototypes received little attention in the previous literature.

    \item Comprehensive experiments on six real-world datasets show that \ti{HETMM} favorably surpasses the state-of-the-art methods, particularly achieving much lower false-positive and missed-detection rates.
 Moreover, using a 60-sheet template set streamlined by \ti{PTS} achieves competitive performance and real-time inference speed (around 26.1 FPS) on a single Quadro 8000 RTX GPU.
 Additionally, \ti{HETMM} is training-free and can be hot-updated by directly inserting novel samples into the template set, which can promptly address some incremental learning issues in industrial manufacturing.
\end{itemize}

\section{Related Works}\label{RW}
\subsection{Anomaly Detection}
The landscape of anomaly detection methods has dramatically evolved over the past decades.
Some surveys \cite{survey2014, survey2019, surveyDL2019} give comprehensive literature reviews.
In this section, we briefly introduce some state-of-the-art approaches.

\noindent\tb{One-Class Classification-based Methods.}
Based on the assumption that the distribution of normal images is significantly different from that of anomalies, Tax and Duin \cite{SVDD} proposed the Support Vector Data Description (SVDD) algorithm to construct a normal decision boundary using hand-crafted support vectors extracted from normal images.
However, subsequent works have proposed the following modifications to improve its performance.
Banerjee \emph{et al.} \cite{fSVDD} introduced a faster version of the original SVDD algorithm.
Ruff \emph{et al.} \cite{DSVDD} replaced the hand-crafted support vectors with deep features to construct a more accurate decision boundary.
Although these modifications have led to better results, the distinction between anomalies and normal samples at the image level is often subtle.
To address this issue, Yi and Yoon \cite{psvdd} proposed constructing patch-level SVDD models that discriminate anomalies based on the patch-level differences between query images and patch-level support vectors.

\noindent\tb{Self-supervised Learning-based Methods.}
To train the anomaly detectors without anomalous samples, self-supervised learning-based methods aim to bypass the need for anomalous samples by a well-designed proxy task.
One notable work is CutPaste \cite{CutPaste}, which generates anomalies by transplanting image patches from one location to another, and DRAEM \cite{DRAEM} leverages the texture dataset DTD \cite{DTD} to synthesize various texture anomalies.
For better anomaly synthesis quality, RealNet \cite{RealNet} employed a diffusion \cite{ddpm} process-based synthesis strategy, capable of generating samples with varying anomaly strengths that mimic the distribution of real anomalous samples. 
While these methods achieve impressive performance, they may yield unsatisfactory outcomes when the anomalous types differ from their expectations.

\noindent\tb{Reconstruction-based Methods.}
Relying on defect-free images to model a latent distribution of normal data, reconstruction-based methods leverage the distribution gap between normal and anomalous patterns to localize anomalous regions.
Bergmann \emph{et al.} \cite{SSIM-AE} have proposed enhancing Convolutional Auto-Encoders (CAEs) \cite{CAE} segmentation results by incorporating structural similarity loss \cite{SSIM}.
In contrast, Schlegl \emph{et al.} \cite{AnoGAN} were the first to use Generative Adversarial Networks (GANs) \cite{GAN} to address the anomaly detection problem.
To improve the effectiveness of \cite{AnoGAN}, Akcay \emph{et al.} \cite{AUROC} and Schlegl \emph{et al.} \cite{F-AnoGan} have proposed an additional encoder to explore the latent representation for better reconstruction quality.
Unlike the above-mentioned methods, \cite{US} reconstructs the normal features selected by teacher networks that are pre-trained on ImageNet \cite{imagenet}.
To incorporate global contexts into local regions, Wang \emph{et al.} \cite{GLFC} have integrated the information obtained from global and local branches simultaneously, while \cite{MKD} aggregates hierarchical information from different resolutions. Bergmann \emph{et al.} \cite{MVTECLOCO} have leveraged regression networks to capture global context and local patch information from a pre-trained feature encoder.
Deng and Li \cite{RD4AD} proposed ``reverse distillation'' that feeds the teacher model's one-class embedding instead of raw images into the student networks to capture better representations of anomalies. 
Batzner \emph{et al.} \cite{EffiAD} proposed a training loss to hinder the student from imitating the teacher feature extractor beyond the normal images, and they further employed an autoencoder to deal with the logical anomalies.
Inspired by the Normalizing Flow (NF) \cite{NF}, DifferNet \cite{DifferNet} first employed it for anomaly detection, while the subsequent advances \cite{csflow,msflow} captured the contexts between hierarchical feature maps for improvements.
Recently, You \emph{et al.} \cite{UniAD} proposed a newly emerging one-for-all scheme, which tried using a unified model to detect anomalies from all the different object classes without finetuning.
For better distinguishing normal and anomalous samples in such a challenging setting, instead of learning the continuous representations, Lu \emph{et al.} \cite{HVQT} preserved the typical normal patterns as discrete iconic prototypes integrated into the Transformer for reconstruction.
However, the above methods' decision boundaries are dominated by the overwhelming number of easy-normal examples.
Thus, they are vulnerable to mistaking the queries near hard-normal examples for anomalous queries.

\noindent\tb{Template-based Methods.}
In addition to the reconstruction-based approaches discussed above, another solution for anomaly detection is constructing a template set with normal features.
Normal and anomalous features, although extracted by the same feature descriptors, significantly differ in feature representations.
Therefore, distinguishing between normal and anomalous patterns is achievable by matching queries with the template set.
Cohen and Hoshen \cite{SPADE} constructed a template set of normal features by extracting the backbone pre-trained on ImageNet \cite{imagenet} for anomaly detection and localization.
At test time, \cite{SPADE} discovers the anomalous patterns with a $K$th Nearest Neighbor ($K$-NN) algorithm on the normal template set.
The inference complexity of \cite{SPADE} is linear to the template set size due to using the $K$-NN algorithm.
Defard \emph{et al.} \cite{Padim} proposed to generate patch embeddings for anomaly localization to reduce the inference complexity of querying the template set.
Reiss \emph{et al.} \cite{PANDA} also proposed an adaptation strategy to finetune the pre-trained features with the corresponding anomaly datasets to obtain high-quality feature representations.
Additionally, PatchCore \cite{patchcore} improves \cite{Padim}'s patch embeddings by extracting neighbor-aware patch-level features and subsampling the template set for speed-up, while Guo \emph{et al.} \cite{THFR} proposed bottleneck compression and template-guided compensation to restore the normal features from anomalies with the guidance of template features.
Bae \emph{et al.} \cite{pni} estimated the normal distribution using conditional probability given by neighborhood features and created a histogram of representative features at each position to leverage the position and neighborhood information.
Similar to the reconstruction-based methods, the queries close to hard-normal examples are also prone to be erroneously identified as anomalies by the above template-based methods, resulting in high false-positive or missed-detection rates.

\subsection{Multi-Prototype Representation}
Multi-prototype representation algorithms aim to select $K$ prototypes to represent the distribution of $N$ samples ($K\le N$).
K-Means-type algorithms \cite{kmeans,fuzzy} attempt to find an optimal partition of the original distribution into $K$ cluster centres.
Liu \emph{et al.} \cite{mpc1} separate the samples into several regions by squared-error clustering and group each high-density region into one prototype.
Nie \emph{et al.} \cite{Nie} select $K$ prototypes using a scalable and parameter-free graph fusion scheme.
Additionally, since coreset \cite{coreset} has been commonly used in K-Means-type algorithms, Roth \emph{et al.} \cite{patchcore} propose a coreset-based selection strategy to find $K$ representatives.
While the multi-prototype representation methods discussed above can effectively cover the distribution of most samples, the coverage of the hard-normal examples is overlooked by those methods.
Consequently, the tiny template set compressed by those methods may struggle to cover the distribution of hard-normal examples.

\section{Methodology}
In this section, we first introduce the preliminary of industrial anomaly detection.
Then we show technical details of the proposed \tb{A}ffine-invariant \tb{T}emplate \tb{M}utual \tb{M}atching (ATMM) and \tb{P}ixel-level \tb{T}emplate \tb{S}election (PTS) modules.
Finally, we describe the overall framework of the proposed \tb{H}ard-normal \tb{E}xample-aware \tb{T}emplate \tb{M}utual \tb{M}atching (HETMM) and how to obtain anomaly detection and localization results.
We provide the notations of this paper in \Cref{table:notations}.

\begin{table}[!t]
  \setlength{\tabcolsep}{0.4mm}
  \caption{The definition of variables in this paper.}
  \label{table:notations}
  \renewcommand{\arraystretch}{1.3}
  \centering
  \footnotesize
  \begin{tabular}{c|m{0.33\textwidth}}
  \toprule
 Symbol                         &Definition\\\midrule 
  $\mathcal{Z}$                     &A set of $N$ collected normal samples.\\
  $z\!\in\!\mathcal{X}^{W\!\times\!H}$          &A normal sample in $\mathcal{Z}$.\\
  $q\!\in\!\mathcal{X}^{W\!\times\!H}$          &A query sample.\\
  $\Phi$                         &A pre-trained network with multiple layers.\\
  $\phi^{(j)}$                      &The $j$-th layer of $\Phi$.\\
  $(x, y)$                        &A pixel index.\\
  $\{\mathcal{T}^{(j)}\}_{j=1}^M$            &The template set with $M$ layer features, where each $\mathcal{T}^{(j)}$ is built by \cref{eq:temp}.\\
  $\mathcal{T}^{(j)}_{x,y}\in\mathbb{R}^{N'\!\times\!C}$ &The template features at the location $(x, y)$, where $N'$ and $C$ denote the size and channels of the template features.\\
  $\mathcal{P}^{\mathcal{T}^{(j)}}_{x,y}\!\!\in\!\mathbb{R}^{n\!\times\!m\!\times\!N'\!\times\!C}$ &An $n\!\times\!m$ patch centred by $\mathcal{T}^{(j)}_{x, y}$ formulated in \cref{eq:PT}.\\
  $\{\mathcal{Q}^{(j)}\}_{j=1}^M$            &A set of query features with $M$ layers, where each $\mathcal{Q}^{(j)}=\phi^{(j)}(q)$ and $\mathcal{Q}^{(j)}_{x,y}\!\in\!\mathbb{R}^{1\times C}$.\\
  $\mathcal{P}^{\mathcal{Q}^{(j)}}_{x,y}\!\!\in\!\mathbb{R}^{n\!\times\!m\!\times\!C}$  &An $n\!\times\!m$ patch centred by $\mathcal{Q}^{(j)}_{x,y}$.\\
  $\mathcal{T}^{(j)*}$                  &The original template set built by all the normal samples, where each $\mathcal{T}^{(j)*}_{x,y}\!\in\!\mathbb{R}^{N\!\times\!C}$. \\
  $\mathcal{T}^{(j)K}$                  &A $K$-sheet tiny set obtained from $\mathcal{T}^{(j)*}$, where each $\mathcal{T}^{(j)K}_{x,y}\!\in\!\mathbb{R}^{K\!\times\!C}$ and $K\leq N$.\\
  $\mathcal{R}=\{r_i\}^L_{i=1}$             &A set of $L$ high-density regions.\\
  $\overrightarrow{S}^{(j)}\in\mathcal{X}^{W'\!\times\!H'}$ &An anomaly map generated by forward \ti{ATM} in \cref{eq:DQT}, where $W'$ and $H'$ denote the width and height of the feature maps.\\
  $\overleftarrow{S}^{(j)}\in\mathcal{X}^{W'\!\times\!H'}$ &An anomaly map generated by backward \ti{ATM} in \cref{eq:DTQ}.\\
  $S^{(j)}\in\mathcal{X}^{W'\!\times\!H'}$        &An anomaly map generated by \ti{ATMM} at the $j$-th layer in \cref{eq:merge}.\\
  $S^\dagger\in\mathcal{X}^{W\!\times\!H}$        &The final anomaly map with the same size of query samples, which can be obtained by \cref{eq:final}.\\
  \bottomrule
  \end{tabular}
\end{table}

\subsection{Preliminary}\label{sec:temp_match}
Industrial anomaly detection aims to identify the unknown defects by the collected $N$ anomaly-free images $\mathcal{Z}$.
Based on the prior hypotheses that most anomalies are significantly different from normal samples, provided a query image $q$ is dissimilar to all the images in $\mathcal{Z}$, it tends to be an anomaly.
Intuitively, let $\Phi$ be a pre-trained network, $q$ can be identified as:
\begin{equation}
 Cat.\!\!=\!\!
  \begin{cases}
 Normal &\!\!\!\!\min\limits_{z\in\mathcal{Z}} \!\left\{d\left(\Phi(q), \Phi(z)\right)\right\}\! < \!\epsilon \\
 Anomalous &\qquad otherwise
  \end{cases}\!\!,
  \label{eq:motivation}
\end{equation}
where $d(\cdot, \cdot)$ and $\epsilon$ denote distance function and anomaly decision threshold, respectively.
For each query, the above template matching attempts to find the corresponding prototypes and then calculates the distance between themselves as an anomaly score.
Existing template-based anomaly detectors \cite{SPADE,Padim,patchcore} propose a two-stage framework: \tb{I)} template generation and \tb{II)} anomaly prediction, to employ template matching for anomaly detection.
In stage \tb{I}, these methods first construct a template set $\{\mathcal{T}^{(j)}\}_{j=1}^M$ using $M$ layers of hierarchical features extracted by a pre-trained network $\Phi$.
Each $\mathcal{T}^{(j)}$ can be obtained by:
\begin{equation}
 \mathcal{T}^{(j)} = \bigcup_{z\in\mathcal{Z}}\phi^{(j)}(z),
  \label{eq:temp}
\end{equation}
where $\phi^{(j)}$ denotes the $j$-layer of $\Phi$.
In stage \tb{II}, for each query image $q$, these methods employ template matching to calculate its anomaly score, and the category can be identified by the threshold determination like \cref{eq:motivation}.
As a result, these template-based methods build a prototype-based decision boundary to distinguish between normal and anomalous queries.

\begin{figure*}[!t]
  \centering
  \includegraphics[width=\textwidth]{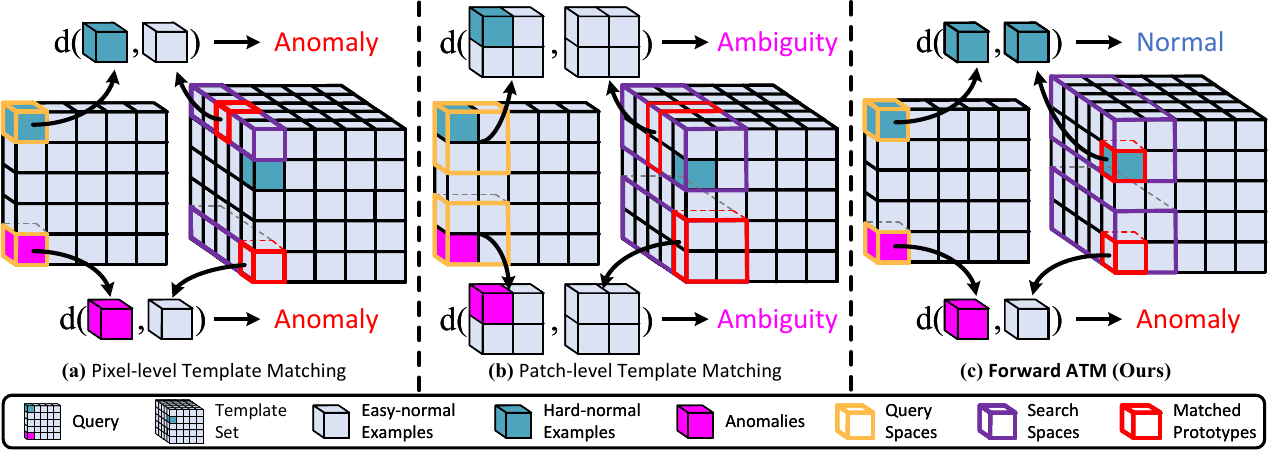}
  \caption{
 Visual examples of different template matching approaches.
 As shown, for each query space (orange frames), pixel-level template matching \tb{(a)} and patch-level template matching \tb{(b)} first search for the similar prototypes within the corresponding pixel- and patch-level search spaces (purple frames) in the template set.
 Then, the corresponding anomaly score is obtained by the distance with its matched prototypes (red frames).
 However, pixel- and patch-level template matching strategies are vulnerable to confusing hard-normal examples (blue cubes) with anomalies (pink cubes).
 Specifically, pixel-level template matching often misses the best-matched prototypes of hard-normal examples due to the slight affine transformations, misclassifying some hard-normal examples as anomalies.
 While patch-level template matching achieves better robustness against affine transformations, the signals of subtle anomalies may be covered by the overwhelming number of easy-normal examples (grey cubes) within the patch, misclassifying some anomalies as normal samples.
 By contrast, the forward \ti{ATM} \tb{(c)} explores the pixel-level prototypes within the corresponding patch-level search space, which can accurately distinguish between hard-normal examples and anomalies.
 }
  \label{fig:HAM}
\end{figure*}

\subsection{Affine-invariant Template Mutual Matching}\label{sec:HAM}
Template-based methods aim to construct a prototype-based decision boundary to detect anomalies from query samples like \cref{eq:motivation}.
Existing template-based methods are based on two template-matching strategies: pixel- and patch-level template matching.
However, the decision boundaries built by those approaches are not robust.
\cref{fig:HAM} depicts the limitations of pixel-level template matching \tb{(a)} and patch-level template matching \tb{(b)}.
As shown, pixel- and patch-level template matching strategies aim to search for similar pixel- and patch-level prototypes in the template set.
However, they are prone to confusing hard-normal examples with anomalies, resulting in many false alarms.
This is because pixel-level template matching may miss the best-matched prototypes due to the affine transformations.
While patch-level template matching has better robustness against affine transformations, it may lead to missed detection due to the distraction of the overwhelming number of easy-normal examples.

To address the above limitations, we propose \ti{ATMM}, a mutual matching between the multi-layer query features $\{\mathcal{Q}^{(j)}\}_{j=1}^M$ and template set, where each $\mathcal{Q}^{(j)}\!\!=\!\!\phi^{(j)}(q)$.
\ti{ATMM} aims to construct a robust prototype-based decision boundary, which comprises forward and backward \tb{A}ffine-invariant \tb{T}emplate \tb{M}atching (ATM) modules.
Specifically, for the $j$-th layer, \ti{ATMM} first normalizes the features of query $\mathcal{Q}^{(j)}$ and template set $\mathcal{T}^{(j)}$ with the $\ell_2$-norm, and then aggregates the anomaly scores towards the forward direction (from $\mathcal{Q}^{(j)}$ to $\mathcal{T}^{(j)}$) and backward direction (from $\mathcal{T}^{(j)}$ to $\mathcal{Q}^{(j)}$), respectively.

\noindent\tb{Forward \ti{ATM}.}
Let the size of template features be $N'$, \ti{i.e.,} containing the features extracted from $N'$ normal samples, $\mathcal{T}^{(j)}_{x,y}\in\mathbb{R}^{N'\!\times\!C}$ denotes the template features at the pixel location $(x, y)$ at the $j$-th layer.
Thus, an $n\!\times\!m$ patch $\mathcal{P}^{\mathcal{T}^{(j)}}_{x,y}\!\in\!\mathbb{R}^{n\!\times\!m\!\times\!N'\!\times\!C}$ centered by $\mathcal{T}^{(j)}_{x,y}$ can be formulated as:
\begin{equation}
 \mathcal{P}^{\mathcal{T}^{(j)}}_{x,y}= \left\{\mathcal{T}^{(j)}_{x+a, y+b}:\lvert a \rvert\le\lfloor\frac{m}{2}\rfloor\land \lvert b\rvert\le\lfloor\frac{n}{2}\rfloor\right\},
  \label{eq:PT}
\end{equation}
where $a,b\in\mathbb{Z}$ and $(x+a, y+b)$ denotes the pixel location within the $n\!\times\!m$ patch.
Given a query feature $\mathcal{Q}^{(j)}_{x,y}\in\mathbb{R}^{1\!\times\!C}$, the forward \ti{ATM} searches for the similar prototypes from $\mathcal{P}^{\mathcal{T}^{(j)}}_{x,y}$ to calculate the corresponding forward anomaly score $\overrightarrow{S}^{(j)}_{x,y}$ as:
\begin{equation}
 \overrightarrow{S}^{(j)}_{x,y} = \!\!\min\limits_{\hat{p}\in\mathcal{P}^{\mathcal{T}^{(j)}}_{x,y}}\!\!\left\{1-\frac{\mathcal{Q}^{(j)}_{x,y}\cdot\hat{p}^{\top}}{\|\mathcal{Q}^{(j)}_{x,y}\|_{2}\cdot\|\hat{p}\|_{2}}\right\},
  \label{eq:DQT}
\end{equation}
where $\hat{p}\in\mathbb{R}^{1\!\times\!C}$.
\cref{fig:HAM} \tb{(c)} shows a visual example of the forward \ti{ATM}.
As shown, the forward \ti{ATM} can construct a robust decision boundary, enabling alleviating the affection brought by affine transformations and easy-normal examples.

\begin{figure}[!t]
  \centering
  \includegraphics[width=0.48\textwidth]{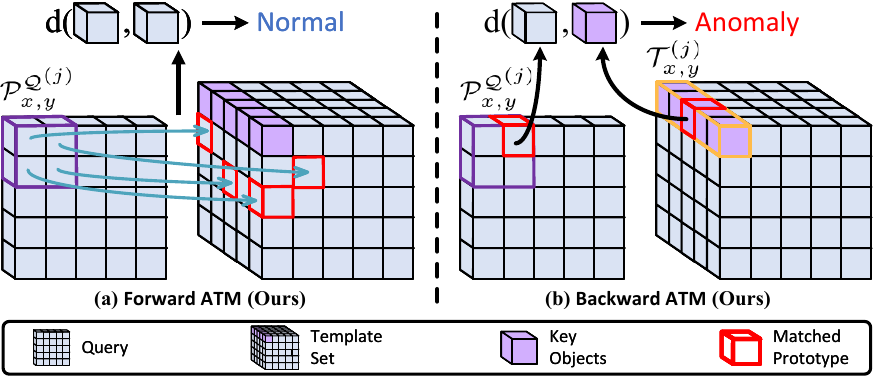}
  \caption{
 For each query object, forward \ti{ATM} \tb{(a)} only explores its similar prototypes (red frames) within the corresponding search space. 
 Therefore, forward \ti{ATM} may misclassify anomalies lacking key objects (purple cubes) in the query patches $\mathcal{P}^{\mathcal{Q}^{(j)}}_{x,y}$ as normal samples.
 On the contrary, for the elements in $\mathcal{T}^{(j)}_{x,y}$, backward \ti{ATM} \tb{(b)} explores similar prototypes within that neighborhood, which can identify whether any key objects are absent, localizing anomalies that are overlooked by forward \textit{ATM}.
 }
  \label{fig:BATM}
\end{figure}

\noindent\tb{Backward \ti{ATM}.}
As reported in \cite{MVTECLOCO}, anomalies can be split into two categories: structural and logical anomalies.
Structural anomalies are invalid objects dissimilar to the template set, while logical ones are valid query objects in invalid locations.
Forward \ti{ATM} can determine whether a query object is valid in the corresponding template patches.
Thus, it can deal with the vast majority of structural anomalies and some logical anomalies.
However, as depicted in \cref{fig:BATM} \tb{(a)}, forward \ti{ATM} may misclassify the anomalous queries lacking key objects as normal ones because it only explores the similar prototypes of query objects, leading to missed detection of some logical anomalies, especially applying forward \ti{ATM} to a large search space.
To address this issue, we further propose backward \ti{ATM}, which employs forward ATM in an inverse direction (\ti{i.e.,} from $\mathcal{T}^{(j)}$ to $\mathcal{Q}^{(j)}$, see \cref{fig:BATM} \tb{(b)}).
For the elements in template feature $\mathcal{T}^{(j)}_{x,y}$, backward \ti{ATM} explores their similar prototypes within the corresponding query patches.
This process enables it to identify whether some key objects are missing in that neighborhood, localizing logical anomalies that are overlooked by forward \textit{ATM}.
Given a query $\mathcal{Q}^{(j)}_{x,y}$, the backward anomaly score $\overleftarrow{S}^{(j)}_{x,y}$ can be calculated as:
\begin{equation} 
 \overleftarrow{S}^{(j)}_{x,y} = \!\!\!\min\limits_{\tilde{p}\in\mathcal{P}^{\mathcal{Q}^{(j)}}_{x,y}}\!\left\{1\!-\!\!\max\limits_{\tilde{t}\in \mathcal{T}^{(j)}_{x,y}}\!\left\{\frac{\tilde{t}\cdot\tilde{p}^{\top}}{\|\tilde{t}\|_{2}\cdot\|\tilde{p}\|_{2}}\right\}\!\right\},
  \label{eq:DTQ}
\end{equation}
where $\mathcal{P}^{\mathcal{Q}^{(j)}}_{x,y}\in\mathbb{R}^{n\!\times\!m\!\times\!C}$ denotes the query $\mathcal{Q}^{(j)}$'s elements within an $n\!\times\!m$ patch centred by the given pixel $(x,y)$, and $\tilde{p}$ and $\tilde{t}$ are belong to $\mathbb{R}^{1\!\times\!C}$.

\begin{figure}[!t]
  \centering
  \includegraphics[width=0.48\textwidth]{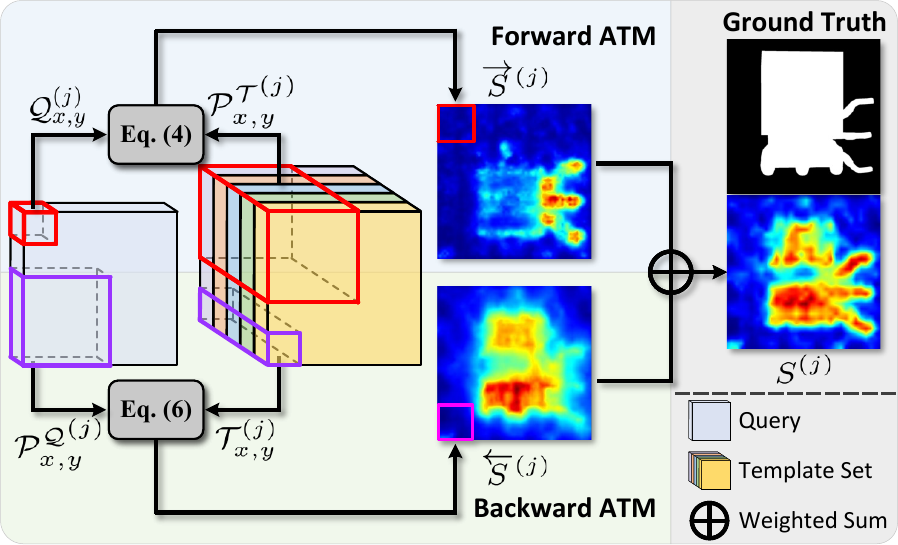}
  \caption{
 The inner structure of \ti{ATMM}.
 For each query image, \ti{ATMM} first generates bi-directional anomaly maps $\overrightarrow{S}^{(j)}$ and $\overleftarrow{S}^{(j)}$ using the forward and backward \ti{ATM} modules, where the anomalous regions in these two maps are complementary to each other.
 The output anomaly map $S^{(j)}$ is the weighted sum of these two maps in \cref{eq:merge}.
 }
  \label{fig:ATMM}
\end{figure}

\begin{figure*}[!t]
  \centering
  \includegraphics[width=\textwidth]{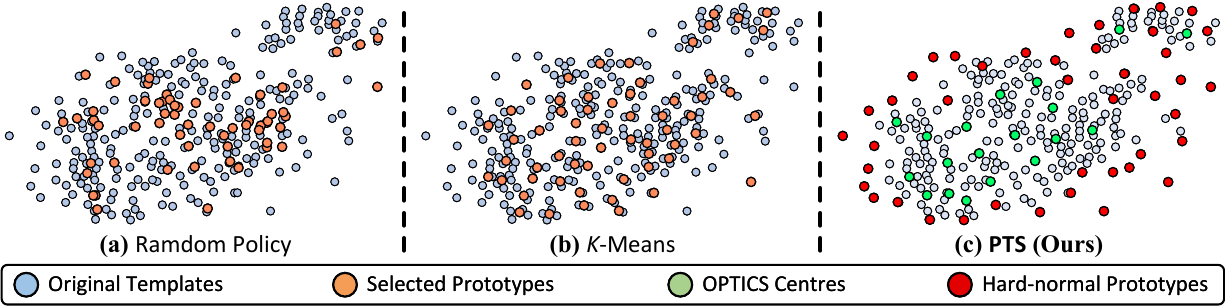}
  \caption{
 The visualization of multi-prototype representation results over the original template set (grey balls) via t-SNE \cite{tsne}.
 Visually, the prototypes selected by random policy \tb{(a)} are rambling, while $K$-Means \tb{(b)} only collect easy-normal prototypes.
 By contrast, \ti{PTS} \tb{(c)} achieves a better distribution coverage of the original template set, collecting OPTICS \cite{OPTICS} centres (green balls) and hard-normal prototypes (red balls) to persist the original decision boundaries.
 }
  \label{fig:PTD}
\end{figure*}

We depict the inner structure of \ti{ATMM} in \cref{fig:ATMM}.
Visually, the anomalous regions recorded in forward and backward anomaly maps (\ti{i.e.,} $\overrightarrow{S}^{(j)}$ and $\overleftarrow{S}^{(j)}$) are complementary to each other, which convincingly confirms the correctness of our motivation.
To achieve more accurate and robust industrial anomaly detection, we merge these bi-directional maps to generate the output maps $S^{(j)}$ of \ti{ATMM} using the weighted sum, which can be formulated as:
\begin{equation}
 S^{(j)}=\alpha\overrightarrow{S}^{(j)}+(1-\alpha)\overleftarrow{S}^{(j)},
  \label{eq:merge}
\end{equation}
where $\alpha$ denotes the ratio to balance the anomalous signals from bi-directions.
Compared with the forward and backward anomaly maps, $S^{(j)}$ achieves better performance in anomaly detection and localization.

\subsection{Pixel-level Template Selection}\label{sec:pts}
For each query, \ti{ATMM} aims to identify its category by measuring the distance between itself and the template set.
Hence, an ideal template set is supposed to represent the distribution of all the normal samples.
However, since industrial images have considerable redundant features, the ideal template set should consist of vast images, leading to considerable memory and computational costs.
Let $\mathcal{T}^{(j)*}$ denote the $j$-th layer features from the original template set, to meet the speed-accuracy demands for industrial production, an ideal solution is to streamline $\mathcal{T}^{(j)*}$ into a tiny one $\mathcal{T}^{(j)K}$ by selecting $K$ prototypes from the $N$ ones ($K\le N$).

There are two frequently-used prototype selection manners: the random policy and $K$-Means \cite{kmeans}.
Random policy aims to randomly select a fraction of data from $\mathcal{\mathcal{T}^{*}}$.
While it can reduce computational costs, its representation capability is neither satisfactory nor stable.
In practical tasks, the cluster centres of $K$-Means are often used for distribution representation.
\cref{fig:PTD} \tb{(a)} and \tb{(b)} visualize the multi-prototype representation results of random policy and the cluster centres of $K$-Means, respectively.
As expected, the distribution of random policy is rambling.
By contrast, the prototypes selected by $K$-Means obtain an acceptable representation performance.
While $K$-Means can cover most of the easy-normal examples, the coverage of the hard-normal ones is likely to be overlooked.
Consequently, the tiny sets obtained by cluster centers struggle to preserve the original decision boundary, which tends to misclassify the hard-normal queries as anomalies, resulting in many false alarms.

In this subsection, we present a novel prototype selection strategy -- \tb{P}ixel-level \tb{T}emplate \tb{S}election (\ti{PTS}) module, which traverses all the pixels to compress the original template set $\mathcal{T}^{(j)*}$ into the tiny one $\mathcal{T}^{(j)K}$ through the sliding windows.
To achieve a comparable distribution and preserve decision boundaries of the original template $\mathcal{T}^{(j)*}$, we argue that not only easy-normal prototypes but some hard-normal ones are also required.
Therefore, the proposed \ti{PTS} aims to select some hard-normal prototypes from $\mathcal{T}^{(j)*}$ after containing sufficient easy-normal prototypes.
Specifically, \ti{PTS} consists of two steps: \tb{I)} easy-normal example-based initialization and \tb{II)} hard-normal prototype selection.
\Cref{alg:PTS} describes the detailed algorithm procedure of the proposed \ti{PTS}.

\begin{figure*}[!t]
  \centering
  \includegraphics[width=\textwidth]{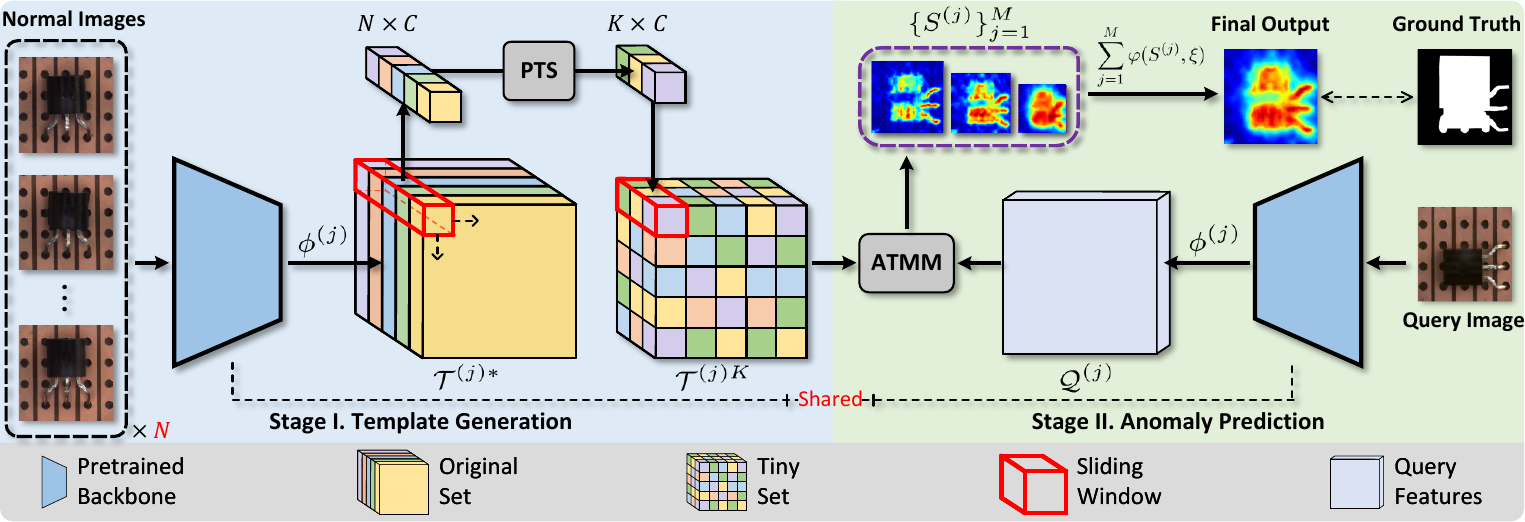}
  \caption{The overall framework of our methods.
 In stage \tb{I}, the original template set $\{\mathcal{T}^{(j)*}\}_{j=1}^M$ is the aggregation of the features extracted by feeding $N$ collected normal images $\mathcal{Z}$ into the pre-trained backbone $\Phi=\{\phi^{(j)}\}^M_{j=1}$ with $M$ layers, where each color on $\mathcal{T}^{(j)*}$ denotes that the feature is extracted from different normal images.
 To streamline $\mathcal{T}^{(j)*}$ into a tiny set $\mathcal{T}^{(j)K}$ with $K$ sheets ($N\ge K$), \ti{PTS} selects $K$ significant prototypes from $\mathcal{T}^{(j)*}$ at each pixel coordinate through the sliding windows.
 In stage \tb{II}, given a query image $q$, we first extract its features $\{Q^{(j)}\}^M_{j=1}$ by the same pre-trained backbone $\Phi$ and then employ \ti{ATMM} to obtain hierarchical anomaly maps $\{S^{(j)}\}^M_{j=1}$, where each $S^{(j)}$ is generated at the $j$-th layer. $S^\dagger$ is obtained using \cref{eq:final} as the final outputs.
 }
  \label{fig:overall}
\end{figure*}

\begin{algorithm2e}[!t]
  \footnotesize
  \SetKwInOut{Input}{input}
  \SetKwInOut{Output}{output}
  \caption{Process of \ti{PTS}.}
  \label{alg:PTS}
  \KwIn{Original template feature $\mathcal{T}^{(j)*}_{x,y}$,\\ \qquad\quad\ The Size of tiny set $K$.}
  \KwOut{Tiny template set pixels $\mathcal{T}^{(j)K}_{x,y}$.}
  \tcc{\emph{Step \tb{I}. Easy-normal Prototype Initialization}}
 Get high-density regions $\mathcal{R}\leftarrow \mathtt{OPTICS}(\mathcal{T}^{(j)*}_{x,y})$\;
  \eIf{$len(\mathcal{R}) > 0$} {
 Tiny template set initialization: $\mathcal{T}^{(j)K}_{x,y}\leftarrow \{\}$\;
    \For{$r_i \in \mathcal{R}$}{
 Get easy-normal prototypes $t_i$ by \cref{eq:optics}\;
 Update tiny set: $\mathcal{T}^{(j)K}_{x,y}\leftarrow \mathcal{T}^{(j)K}_{x,y}\cup \{t_i\}$\;
 }
 } {
 Get global center $t_{g}$ by \cref{eq:gcnt}\;
 Tiny template set initialization: $\mathcal{T}^{(j)K}_{x,y}\leftarrow \{t_{g}\}$\;
 }
  \tcc{\emph{Step \tb{II}. Hard-normal Prototype Selection}}
  \While{$len(\mathcal{T}^{(j)K}_{x,y})<K$}{
 Get hard-normal prototypes $t_{h}$ by \cref{eq:hard}\;
 Update tiny template set: $\mathcal{T}^{(j)K}_{x,y}\leftarrow \mathcal{T}^{(j)K}_{x,y}\cup\{t_{h}\}$\;
 }
\end{algorithm2e}

\noindent\tb{Easy-normal Example-based Initialization.}
Let $(x, y)$ denote the pixel coordinate, for each $\mathcal{T}^{(j)K}_{x,y}$, \ti{PTS} first employs the density clustering method OPTICS \cite{OPTICS} to find the easy-normal examples from $\mathcal{T}^{(j)*}_{x, y}$.
These easy-normal examples are split into several high-density regions $\mathcal{R}=\{r_i\}_{i=1}^{L}$, where $L$ indicates the number of regions.
Since the easy-normal examples are close to each other, the region centre $t_i$ can be regarded as an easy-normal prototype, which can be formulated as:
\begin{equation}
 t_i = \mathop{\arg\max}_{w\in r_i}\left\{\sum_{\hat{w}\in r_i}\frac{w\cdot\hat{w}^{\top}}{\|w\|_{2}\cdot\|\hat{w}\|_{2}}\right\}.
  \label{eq:optics}
\end{equation}
If OPTICS fails to find any high-density regions, which means the features in $\mathcal{T}^{(j)*}_{x, y}$ are scattered far from one another, thereby $\mathcal{T}^{(j)K}_{x,y}$ can be initialized by the global centres $t_g$ as:
\begin{equation}
 t_g = \mathop{\arg\max}\limits_{w\in \mathcal{T}^{(j)*}_{x,y}}\left\{\sum\limits_{\tilde{w}\in \mathcal{T}^{(j)*}_{x,y}}\frac{w\cdot\tilde{w}^{\top}}{\|w\|_{2}\cdot\|\tilde{w}\|_{2}}\right\}.
  \label{eq:gcnt}
\end{equation}
Hence, $\mathcal{T}^{(j)K}_{x,y}$ can be initialized by the aggregation of these selected easy-normal prototypes.

\noindent\tb{Hard-normal Prototype selection.}
Since $\mathcal{T}^{(j)K}_{x, y}$ is initialized by a set of easy-normal prototypes, the easy-normal examples are close to $\mathcal{T}^{(j)K}_{x, y}$ while the hard-normal ones are not.
Let hard-normal prototype $t_h$ be the farthest element from all the elements in the tiny set during each iteration, we can find them based on the maximum value of total distance to all elements in the tiny set $\mathcal{T}^{(j)K}_{x, y}$ as:
\begin{equation}
 t_h \!= \!\!\!\mathop{\arg\max}_{w\in \mathcal{T}^{(j)*}_{x,y}\setminus \mathcal{T}^{(j)K}_{x,y}}\!\left\{\sum_{\tilde{w}\in \mathcal{T}^{(j)K}_{x,y}}\!\left(1\!-\!\frac{w\cdot\tilde{w}^{\top}}{\|w\|_{2}\cdot\|\tilde{w}\|_{2}}\right)\!\right\},
  \label{eq:hard} 
\end{equation}
where the sum operation calculates the total distance to all elements in the tiny set.
This formulation guarantees the selected hard-normal prototypes are not only far away from the easy-normal ones but also far apart from each other, resulting in low redundancy and good coverage of the original distribution.

The distribution coverage of \ti{PTS} is shown in \cref{fig:PTD} \tb{(c)}.
Visually, easy-normal prototypes selected in step \tb{I} cover the most samples in the original distribution.
Moreover, the hard-normal examples scattered at the boundary of the original distribution are also well-covered by the hard-normal prototypes selected in step \tb{II}.
Compared to the results of random policy and $K$-Means, \ti{PTS} achieves better representation performance, especially maintaining the original decision boundaries.

\begin{table*}[!t]
  \caption{
 Quantitative comparisons of start-of-the-arts on the MVTec AD \cite{MVTEC} dataset in terms of the \ti{image-level} \tb{AUROC \%} ($\mathtt{AUC_I}\uparrow$) in this table. 
  \tb{Bold} and \ul{underline} texts indicate the best and second best performance.
 }
  \label{table:IAD}
  \renewcommand{\arraystretch}{1.2}
  \centering
  \tiny
  \setlength{\tabcolsep}{0.5mm}
  \setlength{\belowbottomsep}{3pt}
  \begin{tabular}{c|c|cccccccccc|cc}
  \bottomrule
  \multicolumn{2}{c|}{\multirow{2}{*}{\texttt{Category}}} 
                       &CSFlow\cite{csflow} &PatchCore\cite{patchcore} &RD4AD\cite{RD4AD} &UniAD\cite{UniAD} &PNI\cite{pni} &THFR\cite{THFR} &HVQT\cite{HVQT} &EffiAD\cite{EffiAD}&MSFlow\cite{msflow} &RealNet\cite{RealNet}&\multicolumn{2}{c}{\tb{HETMM}}\\
  \multicolumn{2}{c|}{}           &(WACV'22)      &(CVPR'22)         &(CVPR'22)     &(NIPS'22)     &(ICCV'23)   &(ICCV'23)    &(NIPS'23)    &(WACV'24)     &(TNNLS'24)     &(CVPR'24)      &(\ti{ALL})   &(\ti{60 sheets}) \\\midrule
  \multirow{6}{*}{\rb{\texttt{Textures}}}           
  &\texttt{Carpet}             &\tb{100.0}     &98.0           &98.9       &\ul{99.9}     &\tb{100.0}  &99.8      &\ul{99.9}    &99.3        &\ul{99.9}      &99.8         &\tb{100.0}   &99.8       \\
  &\texttt{Grid}              &99.0        &98.6           &\tb{100.0}    &98.5       &98.4     &\tb{100.0}   &97.0      &\ul{99.9}     &99.5        &\tb{100.0}      &\tb{100.0}   &\ul{99.9}    \\
  &\texttt{Leather}             &\tb{100.0}     &\tb{100.0}        &\tb{100.0}    &\tb{100.0}    &\tb{100.0}  &\tb{100.0}   &\tb{100.0}   &\tb{100.0}     &\tb{100.0}     &\tb{100.0}      &\tb{100.0}   &\tb{100.0}    \\
  &\texttt{Tile}              &\tb{100.0}     &99.4           &99.3       &99.0       &\tb{100.0}  &99.3      &99.2      &\tb{100.0}     &\ul{99.9}      &\tb{100.0}      &\tb{100.0}   &\ul{99.9}    \\
  &\texttt{Wood}              &\tb{100.0}     &99.2           &99.2       &97.9       &99.6     &99.2      &97.2      &99.5        &\tb{99.8}      &99.2         &\ul{99.7}   &\ul{99.7}    \\
  \cline{2-14}\noalign{\smallskip}             
  &\texttt{Average}             &\ul{99.8}      &99.1           &99.5       &99.1       &99.6     &99.7      &98.7      &99.7        &\ul{99.8}      &\ul{99.8}      &\tb{99.9}   &\tb{99.9}    \\\midrule
  \multirow{11}{*}{\rb{\texttt{Objects}}}               
  &\texttt{Bottle}             &\ul{99.8}      &\tb{100.0}        &\tb{100.0}    &\tb{100.0}    &\tb{100.0}  &\tb{100.0}   &\tb{100.0}   &\tb{100.0}     &\tb{100.0}     &\tb{100.0}      &\tb{100.0}   &\tb{100.0}    \\
  &\texttt{Cable}              &99.1        &99.3           &95.0       &97.6       &\ul{99.8}   &99.2      &99.0      &95.2        &99.3        &99.2         &\tb{100.0}   &\tb{100.0}    \\
  &\texttt{Capsule}             &97.1        &98.0           &96.3       &85.3       &\ul{99.7}   &97.5      &95.4      &97.9        &99.3        &99.5         &\tb{100.0}   &\tb{100.0}    \\
  &\texttt{Hazelnut}            &99.6        &\tb{100.0}        &\ul{99.9}     &\ul{99.9}     &\tb{100.0}  &\tb{100.0}   &\tb{100.0}   &99.4        &\tb{100.0}     &\tb{100.0}      &\tb{100.0}   &\tb{100.0}    \\
  &\texttt{Metal nut}            &99.1        &99.7           &\tb{100.0}    &99.0       &\tb{100.0}  &\tb{100.0}   &\ul{99.9}    &99.6        &\tb{100.0}     &99.8         &\tb{100.0}   &\tb{100.0}    \\
  &\texttt{Pill}              &98.6        &97.0           &96.6       &88.3       &96.7     &97.8      &95.8      &98.6        &\ul{98.7}      &\tb{99.1}      &98.2      &97.5       \\
  &\texttt{Screw}              &97.6        &96.4           &97.0       &91.9       &\tb{99.5}   &97.1      &95.6      &97.0        &97.6        &98.8         &\ul{99.2}   &95.8       \\
  &\texttt{Toothbrush}           &91.9        &\tb{100.0}        &99.5       &95.0       &\ul{99.7}   &\tb{100.0}   &93.6      &\tb{100.0}     &\tb{100.0}     &99.4         &\tb{100.0}   &\tb{100.0}    \\
  &\texttt{Transistor}           &99.3        &\ul{99.9}         &96.7       &\tb{100.0}    &\tb{100.0}  &99.7      &99.7      &\ul{99.9}     &\tb{100.0}     &\tb{100.0}      &\tb{100.0}   &\tb{100.0}    \\
  &\texttt{Zipper}             &99.7        &99.2           &98.5       &96.7       &\tb{99.9}   &97.7      &97.9      &99.7        &\ul{99.8}      &\ul{99.8}      &\ul{99.8}   &99.2       \\
  \cline{2-14}\noalign{\smallskip}             
  &\texttt{Average}             &98.2        &99.0           &98.0       &95.4       &99.5     &98.9      &97.7      &98.7        &99.5        &\ul{99.6}      &\tb{99.7}   &99.2       \\\midrule 
  \multicolumn{2}{c|}{\tb{Average}}                 
                       &98.7        &99.1           &98.5       &96.6       &99.5     &99.2      &98.0      &99.1        &\ul{99.6}      &\ul{99.6}      &\tb{99.8}   &99.5       \\\bottomrule 
  \end{tabular}
\end{table*}

\subsection{Overall Framework}
The overall framework of our methods is illustrated in \cref{fig:overall}.
Similar to the existing template-based methods \cite{SPADE,patchcore}, the proposed framework also comprises two stages: \tb{I)} template generation, \tb{II)} anomaly prediction, as reported in Section \ref{sec:temp_match}.
In stage \tb{I}, given the original template set $\{\mathcal{T}^{(j)*}\}^M_{j=1}$, we employ the \ti{PTS} to compress each $\mathcal{T}^{(j)*}$ into a $K$-sheet one $\mathcal{T}^{(j)K}$ to build the tiny set $\{\mathcal{T}^{(j)K}\}^M_{j=1}$.
In stage \tb{II}, for each query image $q$, we first extract its features $\{\mathcal{Q}^{(j)}\}^M_{j=1}$ by the same pre-trained model $\Phi$.
Then, the hierarchical anomaly maps $\{S^{(j)}\}^M_{j=1}$ can be obtained by applying \ti{ATMM} to the query and template features extracted from each layer in $\{\phi^{(j)}\}^M_{j=1}$, which can capture the contextual information from multiple resolutions for more accurate anomaly detection and localization.
The final outputs $S^\dagger$ is calculated as:
\begin{equation}
 S^\dagger = \sum_{j=1}^M\varphi(S^{(j)},\xi),
  \label{eq:final}
\end{equation}
where $\varphi(\cdot, \xi)$ denotes the rescale operation to upsample the anomaly maps with the $\xi$ resolutions, while the default value of $\xi$ equals the resolution of the query image $q$.

\begin{table*}[!t]
  \caption{
 Quantitative comparisons of start-of-the-arts on the MVTec AD \cite{MVTEC} dataset in terms of the \ti{pixel-level} \tb{AUROC \%} ($\mathtt{AUC_P}\uparrow$) in this table. 
  \tb{Bold} and \ul{underline} texts indicate the best and second best performance.
 }
  \label{table:PAL}
  \renewcommand{\arraystretch}{1.2}
  \centering
  \tiny
  \setlength{\tabcolsep}{1mm}
  \setlength{\belowbottomsep}{3pt}
  \begin{tabular}{c|c|ccccccccc|cc}
  \bottomrule
  \multicolumn{2}{c|}{\multirow{2}{*}{\texttt{Category}}} 
                        &PatchCore\cite{patchcore} &RD4AD\cite{RD4AD} &UniAD\cite{UniAD} &PNI\cite{pni} &THFR\cite{THFR} &HVQT\cite{HVQT} &EffiAD\cite{EffiAD}&MSFlow\cite{msflow} &RealNet\cite{RealNet}&\multicolumn{2}{c}{\tb{HETMM}} \\
  \multicolumn{2}{c|}{}            &(CVPR'22)         &(CVPR'22)     &(NIPS'22)     &(ICCV'23)   &(ICCV'23)    &(NIPS'23)    &(WACV'24)     &(TNNLS'24)     &(CVPR'24)      &(\ti{ALL})   &(\ti{60 sheets}) \\\midrule
  \multirow{6}{*}{\rb{\texttt{Textures}}}  
  &\texttt{Carpet}              &98.9           &98.9       &98.5       &\ul{99.4}   &99.2      &98.7      &96.3        &99.3        &99.2         &\tb{99.5}   &99.2     \\
  &\texttt{Grid}               &98.6           &\ul{99.3}     &96.5       &99.2     &99.3      &97.0      &94.1        &99.4        &\ul{99.5}       &\tb{99.6}   &98.8     \\
  &\texttt{Leather}              &99.3           &99.4       &98.8       &99.6     &99.4      &98.8      &97.7        &\ul{99.7}     &\tb{99.8}       &\tb{99.8}   &\ul{99.7}  \\
  &\texttt{Tile}               &96.1           &95.6       &91.8       &\ul{98.4}   &95.5      &92.2      &91.5        &98.1        &\tb{99.4}       &97.6      &97.3     \\
  &\texttt{Wood}               &95.1           &95.3       &93.2       &\ul{97.0}   &95.3      &92.4      &90.9        &96.3        &\tb{98.2}       &96.8      &96.4     \\
  \cline{2-13}\noalign{\smallskip}             
  &\texttt{Average}              &97.6           &97.7       &95.8       &\ul{98.7}   &97.7      &95.8      &94.1        &98.6        &\tb{99.2}       &\ul{98.7}   &98.3     \\\midrule
  \multirow{11}{*}{\rb{\texttt{Objects}}}               
  &\texttt{Bottle}              &98.5           &98.7       &98.1       &98.9     &98.9      &98.3      &98.7        &98.7        &\tb{99.3}       &\ul{99.2}   &98.8     \\
  &\texttt{Cable}               &98.2           &97.4       &97.3       &\tb{99.1}   &98.5      &98.1      &\ul{98.8}     &98.4        &98.1         &\ul{98.8}   &98.4     \\
  &\texttt{Capsule}              &98.8           &98.7       &98.5       &\tb{99.3}   &98.7      &98.8      &\ul{99.2}     &\ul{99.2}     &\tb{99.3}       &99.0      &98.5     \\
  &\texttt{Hazelnut}             &98.6           &98.9       &98.1       &99.4     &99.2      &98.8      &98.8        &98.6        &\tb{99.7}       &\ul{99.5}   &99.4     \\
  &\texttt{Metal nut}             &98.4           &97.3       &94.8       &\tb{99.3}   &97.4      &96.3      &98.5        &99.3        &98.6         &\ul{98.7}   &97.5     \\
  &\texttt{Pill}               &97.1           &98.2       &95.0       &\tb{99.0}   &98.0      &97.1      &98.7        &\ul{98.9}     &\tb{99.0}       &98.7      &97.8     \\
  &\texttt{Screw}               &99.2           &\tb{99.6}     &98.3       &\tb{99.6}   &\ul{99.5}    &98.9      &98.7        &99.3        &\ul{99.5}       &\tb{99.6}   &99.3     \\
  &\texttt{Toothbrush}            &98.5           &99.1       &98.4       &99.1     &99.2      &98.6      &97.7        &98.7        &98.7         &\tb{99.5}   &\ul{99.4}  \\
  &\texttt{Transistor}            &94.9           &92.5       &97.9       &98.0     &95.9      &97.9      &97.2        &95.1        &98.0         &\tb{99.2}   &\ul{98.4}  \\
  &\texttt{Zipper}              &98.8           &98.2       &96.8       &\tb{99.4}   &98.7      &97.5      &96.3        &99.0        &\ul{99.2}       &99.0      &98.1     \\
  \cline{2-13}\noalign{\smallskip}             
  &\texttt{Average}              &98.1           &97.9       &97.3       &\tb{99.1}   &98.4      &98.0      &98.3        &98.5        &\ul{98.9}       &\tb{99.1}   &98.6     \\\midrule 
  \multicolumn{2}{c|}{\tb{Average}}             
                        &98.1           &97.8       &96.8       &\tb{99.0}   &98.2      &97.3      &96.9        &98.5        &\tb{99.0}       &\tb{99.0}   &\ul{98.7}   \\\bottomrule 
  \end{tabular}
\end{table*}

\subsection{Detection and Localization}
Given the final anomaly map $S^\dagger$ by integrating anomaly maps $S^{(j)}$ at each layer in \cref{eq:final}, we can obtain anomaly detection and localization results by some frequently-used post-processing techniques.
Following the previous works \cite{patchcore,SPADE,Padim}, the anomaly detection results $S^{D}$ are calculated by:
\begin{equation}
 S^{D} = \mathtt{max}(\mathcal{G}(S^\dagger, \sigma)),
\end{equation}
where $\mathcal{G}(\cdot, \sigma)$ denotes the Gaussian blur filter with $\sigma=6.8$ as default, while $\mathtt{max}$ indicates the global maximum operation.
The anomaly localization result $S^{L}$ can be directly obtained by employing the 0-1 normalization operation to the anomaly map $S^\dagger$ without demanding other post-processing techniques.


\section{Experiments \& Analysis}
In this section, we conduct extensive experiments and in-depth analyses to demonstrate the superiority of the proposed methods in anomaly detection and localization.
Following the one-class classification protocols, we only leverage normal images to detect and localize unknown anomalies.

\subsection{Experimental Details}

\noindent\tb{Datasets.}\label{sec:dataset}
We comprehensively compare our methods and the state-of-the-art approaches on the following six real-world datasets (five industrial datasets and one non-industrial dataset).
The comprehensive evaluation of the MVTec AD \cite{MVTEC} dataset is reported in \cref{EXP:MVTec}, while the other five datasets are evaluated in \cref{EXP:Others}.

\tb{MVTec Anomaly Detection} (MVTec AD) benchmark \cite{MVTEC} consists of 15 categories with a total of 5354 high-resolution images, of which 3629 anomaly-free images are used for training and the rest 1725 images, including normal and anomalous ones for testing.
Each category owns 60 to 390 anomaly-free training images. 
Both image- and pixel-level annotations are provided in its test set.
As a real-world industrial dataset, most of our experiments are conducted on the MVTec AD to evaluate industrial anomaly detection and localization performance.

\tb{Visual Anomaly} (VisA) dataset \cite{VisA} consists of 10821 high-resolution color images, including 9621 normal and 1200 anomalous samples.
It provides image- and pixel-level labels, covering 12 objects in 3 domains, making it the largest and most challenging public dataset for anomaly classification and segmentation.
In addition to the standard 1-class training setup, the VisA \cite{VisA} dataset also supports 2-class training setups with 5/10/high-shot.

\tb{Magnetic Tile Defects} (MTD) dataset \cite{MTD}, a specialized real-world industrial dataset for the detection of magnetic tile defects.
MTD contains 952 defect-free images, while the rest 395 defective magnetic tile images with various illumination levels split into $5$ defect types.
Following the setting reported in \cite{DifferNet}, we select 80\% normal magnetic tile images as anomaly-free training data, the rest 20\% images for the test.
The experiments performed on the MTD can demonstrate the models' capability of detecting anomalies in various illuminative environments.

\begin{table*}[!t]
  \caption{
 Quantitative comparisons of start-of-the-arts on the MVTec AD \cite{MVTEC} dataset in terms of the \ti{pixel-level} \tb{PRO \%} ($\mathtt{PRO}\uparrow$) in this table.
  \tb{Bold} and \ul{underline} texts indicate the best and second best performance.
 }
  \label{table:PRO}
  \centering
  \tiny
  \setlength{\tabcolsep}{1mm}
  \setlength{\belowbottomsep}{3pt}
  \begin{tabular}{c|c|ccccccccc|cc}
  \bottomrule
  \multicolumn{2}{c|}{\multirow{2}{*}{\texttt{Category}}} 
                       &PatchCore\cite{patchcore} &RD4AD\cite{RD4AD} &UniAD\cite{UniAD} &PNI\cite{pni} &THFR\cite{THFR} &HVQT\cite{HVQT} &EffiAD\cite{EffiAD}&MSFlow\cite{msflow} &RealNet\cite{RealNet}&\multicolumn{2}{c}{\tb{HETMM}} \\
  \multicolumn{2}{c|}{}           &(CVPR'22)         &(CVPR'22)     &(NIPS'22)     &(ICCV'23)   &(ICCV'23)   &(NIPS'23)    &(WACV'24)     &(TNNLS'24)     &(CVPR'24)      &(\ti{ALL})   &(\ti{60 sheets}) \\\midrule
  \multirow{6}{*}{\rb{\texttt{Textures}}}      
  &\texttt{Carpet}              &96.5           &97.0       &94.2       &97.6     &\ul{97.7}    &95.1      &92.7        &\tb{99.3}     &96.4         &97.5      &97.1     \\
  &\texttt{Grid}               &96.1           &97.6       &92.1       &94.3     &\ul{97.7}    &90.0      &88.9        &\tb{98.6}     &97.3         &97.5      &96.6     \\
  &\texttt{Leather}             &98.9           &99.1       &97.2       &98.3     &\ul{99.2}    &97.7      &98.3        &\tb{99.9}     &96.2         &98.8      &98.7     \\
  &\texttt{Tile}               &88.3           &90.6       &82.6       &\ul{94.7}   &90.8      &85.3      &85.7        &93.9        &\tb{97.7}      &91.7      &90.6     \\
  &\texttt{Wood}               &89.5           &90.9       &87.2       &92.5     &93.3      &89.1      &90.2        &93.5        &90.5         &\tb{95.2}   &\ul{95.0}  \\
  \cline{2-13}\noalign{\smallskip}             
  &\texttt{Average}             &93.9           &95.0       &90.6       &95.5     &95.7      &91.5      &91.2        &\tb{97.0}     &95.6         &\ul{96.2}   &95.6     \\\midrule
  \multirow{11}{*}{\rb{\texttt{Objects}}}         
  &\texttt{Bottle}              &95.9           &96.6       &93.6       &95.9     &\ul{97.2}    &94.5      &95.7        &\tb{98.4}     &95.6         &96.5      &96.4     \\
  &\texttt{Cable}              &91.6           &91.0       &86.7       &\tb{98.9}   &94.8      &89.0      &92.5        &92.7        &93.4         &\ul{95.4}   &95.0     \\
  &\texttt{Capsule}             &95.5           &95.8       &90.3       &95.6     &95.9      &91.0      &\tb{97.6}     &\ul{97.5}     &84.5         &96.8      &96.5     \\
  &\texttt{Hazelnut}             &93.8           &95.5       &93.8       &96.9     &96.2      &93.1      &95.7        &\tb{97.0}     &93.1         &\ul{96.9}   &96.8     \\
  &\texttt{Metal nut}            &91.2           &92.3       &87.2       &\tb{95.9}   &90.5      &90.9      &94.4        &92.9        &94.4         &\ul{95.3}   &94.2     \\
  &\texttt{Pill}               &92.9           &96.4       &95.3       &96.7     &96.4      &95.4      &96.1        &95.2        &91.0         &\tb{97.2}   &\ul{97.0}  \\
  &\texttt{Screw}              &97.1           &\tb{98.2}     &95.3       &97.2     &\tb{98.2}    &94.4      &96.4        &96.0        &87.9         &\ul{98.1}   &96.7     \\
  &\texttt{Toothbrush}            &90.2           &94.5       &88.0       &92.7     &\ul{94.7}    &89.2      &93.3        &90.8        &91.6         &\tb{94.9}   &\tb{94.9}  \\
  &\texttt{Transistor}            &81.2           &78.0       &93.6       &\ul{96.2}   &85.9      &95.5      &91.2        &92.7        &92.9         &\tb{96.5}   &95.6     \\
  &\texttt{Zipper}              &97.0           &95.4       &93.0       &\ul{97.3}   &96.6      &91.9      &93.4        &\tb{98.5}     &93.4         &96.7      &96.0     \\
  \cline{2-13}\noalign{\smallskip}             
  &\texttt{Average}             &92.7           &93.4       &91.7       &\ul{96.3}   &94.6      &91.5      &94.6        &95.2        &91.8         &\tb{96.5}   &95.9     \\\midrule 
  \multicolumn{2}{c|}{\tb{Average}}            
                       &93.1           &93.9       &91.3       &\ul{96.0}   &95.0      &91.5      &93.5        &95.8        &93.1         &\tb{96.4}   &95.8     \\\bottomrule 
  \end{tabular}
\end{table*}

\begin{figure*}[!t]
  \centering
    \includegraphics[width=\textwidth]{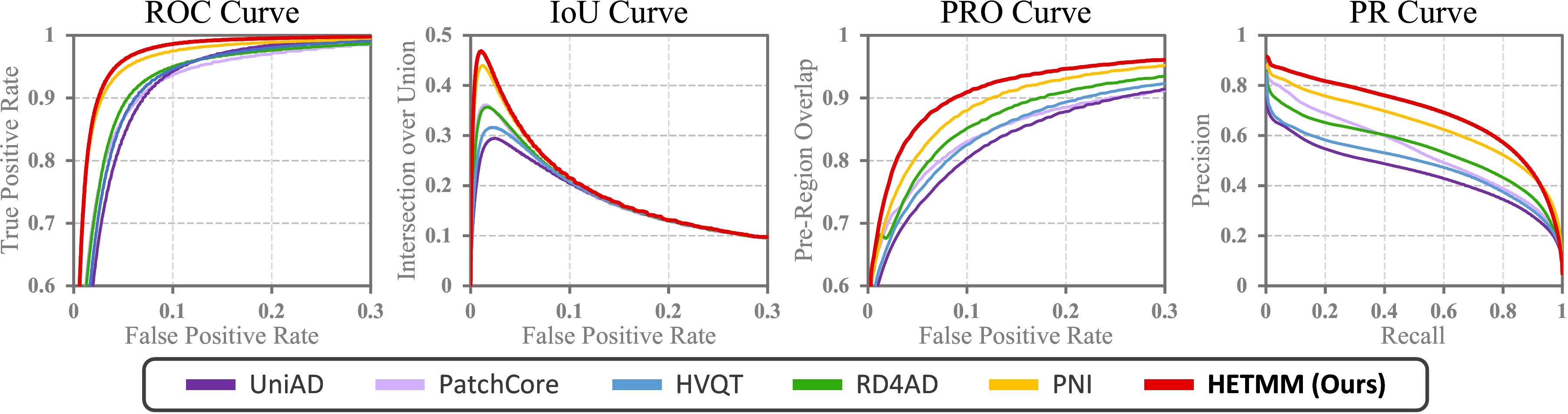}
  \caption{Performance curves of the proposed \tb{HETMM} and state-of-the-art methods on the MVTec AD \cite{MVTEC} dataset.
 }
  \label{fig:curves}
\end{figure*}

\tb{MVTec Logical Constraints Anomaly Detection} (MVTec LOCO) \cite{MVTECLOCO} is the latest anomaly detection dataset with 3644 real-world industrial images, of which 1772 anomaly-free images for training, 304 anomaly-free images for validation and the rest (575 anomaly-free and 993 defective images) belong to the test set.
It consists of 5 object categories from industrial inspection scenarios.
Unlike the one-class classification settings, MVTec LOCO employs a two-class classification, classifying various defects into two categories: structural and logical anomalies.
Compared to the existing anomaly detection dataset, the performance evaluated on the MVTec LOCO places more emphasis on global contexts.

\tb{MVTec Structural and Logical Anomaly Detection} (MVTec SL) \cite{MVTECLOCO} is a dataset that employs the above two-class classification settings on the original MVTec AD.
Only three categories contain 37 logical anomalous images in 1258 anomaly images, while the rest are structural ones.

\tb{Mini Shanghai Tech Campus} (mSTC) \cite{mSTC} dataset is an abnormal event dataset.
Following the settings reported in \cite{CAVGA-R,Padim,patchcore}, we subsample the original Shanghai Tech Campus (STC) dataset by extracting every fifth training and test video frame.
In 12 scenes, mSTC's training data are the normal pedestrian behaviors, while the abnormal behaviors, such as fighting or cycling, belong to its test set.
Compared to the above industrial datasets, localizing the abnormal events is a long-range spatiotemporal challenge.


\begin{figure*}[!t]
  \centering
    \includegraphics[width=\textwidth]{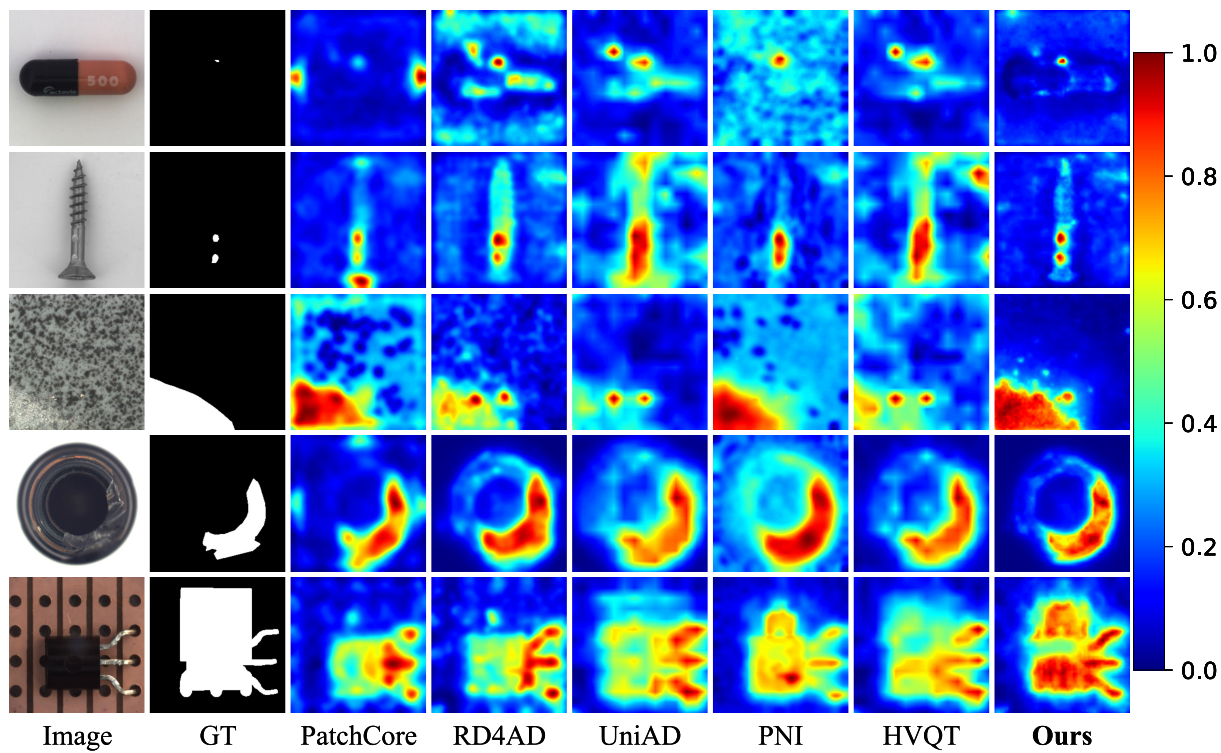}
  \caption{Visual comparisons between the proposed \tb{HETMM} and other state-of-the-arts on the MVTec AD \cite{MVTEC} dataset.
 }
  \label{fig:visual}
\end{figure*}

\noindent\tb{Evaluation Metrics.}\label{sec:metrics}
Following the settings of recent anomaly detection methods \cite{US,patchcore}, we employ the Area Under the Receiver Operating Characteristic Curve (AUROC) and the Per-Region Overlap (PRO) \cite{MVTEC} metrics for evaluations.
The image- and pixel-level AUROC are employed to measure the performance of image-level anomaly detection and pixel-level anomaly localization, respectively.
In addition, we also use the PRO to evaluate the anomaly localization quality.
Let $C_{i,k}$ denote the anomalous regions for a connected component $k$ in the ground truth image $i$ and $P_i$ denote the predicted anomalous regions for a threshold $t$.
Then, the PRO can be formulated as:
\begin{equation}
 PRO=\frac{1}{N}\sum_i\sum_{k}\frac{\vert P_i\cap C_{i,k}\vert}{\vert C_{i,k}\vert },
\end{equation}
where $N$ is the number of ground truth components.
As proposed in \cite{US}, We average the PRO score with false-positive rates below 30\%.
Therefore, a higher PRO score indicates that anomalous regions are well-localized with lower false-positive rates.

\noindent\tb{Implementation Details.}
Our experiments are based on Pytorch \cite{Pytorch} framework and run with a single Quadro 8000 RTX.
$\phi$ here represents the WResNet101 \cite{wresnet} backbone pre-trained on ImageNet \cite{imagenet} without fine-tuning.
The OPTICS algorithm we used is re-implemented by scikit-learn \cite{scikit-learn}.
Following the settings in \cite{CutPaste}, we rescale all the images into 256$\times$256 in our experiments.
Since class-specific data augmentations require prior knowledge of inspected items, we do not use any data augmentation techniques for fair comparisons.

\subsection{Evaluations on the MVTec AD} \label{EXP:MVTec}
In this subsection, we comprehensively compare the proposed \ti{HETMM} with 13 state-of-the-art methods, especially the very recent methods (EffiAD \cite{EffiAD}, MSFlow \cite{msflow}, and RealNet \cite{RealNet}), for anomaly detection and localization on the MVTec AD dataset in terms of the image-level AUROC ($\mathtt{AUC_I}$), pixel-level AUROC ($\mathtt{AUC_P}$) and PRO (\texttt{PRO}). The results of MSFlow \cite{msflow} is the average value of five runs, while the performance of other methods is taken from the results reported in their papers.
We select two different template sets for evaluation: the original template set and the 60-sheet tiny set compressed by \ti{PTS}, which is denoted as \ti{ALL} and \ti{60 sheets}, respectively.

\noindent\tb{Image-level Anomaly Detection.}
\Cref{table:IAD} reports the quantitative evaluation results of image-level AUROC.
As demonstrated, the average performance of the original \ti{HETMM} (\tb{99.8\%} $\mathtt{AUC_I}$) favorably surpasses all the competitors with consistent outperformance in all categories, suggesting \ti{HETMM} has good robustness in the detection of various anomalous types.
In addition, 60-sheet \ti{HETMM} yields comparable performance (\tb{99.5\%} $\mathtt{AUC_I}$) to the recent benchmarks: PNI \cite{pni}, MSFlow \cite{msflow}, and RealNet \cite{RealNet}, and surpasses the rest compared methods by over \tb{0.3\%} $\mathtt{AUC_I}$, which reveals \ti{PTS} significantly reduces the memory and computational costs with yielding superior performance.

\begin{table*}[!t]
  \caption{
 Quantitative comparisons of start-of-the-arts on the VisA \cite{VisA} dataset in terms of image-level \tb{AUROC \%} ($\mathtt{AUC_I}\uparrow$) and pixel-level \tb{AUROC \%} ($\mathtt{AUC_P}\uparrow$) in this table.
  \tb{Bold} and \ul{underline} texts indicate the best and second best performance.
 }
  \label{table:VisA}
  \centering
  \footnotesize
  \begin{tabular}{l|ccccccccccccccccc}
  \toprule
  \multirow{4}{*}{\tb{Category}} 
            &\multicolumn{2}{c}{PatchCore\cite{patchcore}}&&\multicolumn{2}{c}{UniAD\cite{UniAD}}&&\multicolumn{2}{c}{HVQT\cite{HVQT}}&&\multicolumn{2}{c}{EffiAD\cite{EffiAD}}&&\multicolumn{2}{c}{RealNet\cite{RealNet}}&&\multicolumn{2}{c}{\multirow{2}{*}{\tb{HETMM}}}\\
            &\multicolumn{2}{c}{(CVPR'22)}        &&\multicolumn{2}{c}{(NIPS'22)}    &&\multicolumn{2}{c}{NIPS'23}    &&\multicolumn{2}{c}{(WACV'24)}     &&\multicolumn{2}{c}{(CVPR'24)}      &&        &               \\
            \cmidrule{2-3}\cmidrule{5-6}\cmidrule{8-9}\cmidrule{11-12}\cmidrule{14-15}\cmidrule{17-18}
            &$\mathtt{AUC_I}$&$\mathtt{AUC_P}$      &&$\mathtt{AUC_I}$&$\mathtt{AUC_P}$  &&$\mathtt{AUC_I}$&$\mathtt{AUC_P}$ &&$\mathtt{AUC_I}$&$\mathtt{AUC_P}$   &&$\mathtt{AUC_I}$&$\mathtt{AUC_P}$    &&$\mathtt{AUC_I}$&$\mathtt{AUC_P}$       \\\midrule
  \texttt{Candle}   &\ul{98.6}    &\tb{99.5}          &&96.8      &\ul{99.2}      &&96.8      &99.2       &&98.4      &99.1          &&96.1      &99.1          &&\tb{98.8}    &99.0             \\
  \texttt{Capsules}  &81.6      &\tb{99.5}          &&72.0      &98.3        &&77.1      &99.0       &&\tb{93.5}    &98.2          &&\ul{93.2}    &98.7          &&93.1      &\ul{99.4}           \\
  \texttt{Cashew}   &97.3      &\ul{98.9}          &&92.4      &98.7        &&94.9      &\tb{99.2}    &&97.2      &\tb{99.2}       &&\ul{97.8}    &98.3          &&\tb{99.0}    &98.6             \\
  \texttt{Chewinggum} &99.1      &99.1            &&\ul{99.4}    &\ul{99.2}      &&\ul{99.4}    &98.8       &&\tb{99.9}    &\ul{99.2}       &&\tb{99.9}    &\tb{99.8}        &&\ul{99.4}    &98.8             \\
  \texttt{Fryum}   &96.2      &93.8            &&89.8      &\tb{97.7}      &&90.4      &\tb{97.7}    &&96.5      &96.5          &&\tb{97.1}    &96.2          &&\ul{97.0}    &\ul{97.2}           \\
  \texttt{Macaroni1} &97.5      &\ul{99.8}          &&92.2      &99.3        &&93.1      &99.4       &&\ul{99.4}    &\tb{99.9}       &&\tb{99.8}    &\tb{99.9}        &&98.7      &99.7             \\
  \texttt{Macaroni2} &78.1      &99.1            &&85.9      &98.0        &&86.2      &98.5       &&\tb{96.7}    &\ul{99.8}       &&\ul{95.2}    &99.6          &&\ul{95.2}    &\tb{99.9}           \\
  \texttt{Pcb1}    &\ul{98.5}    &\tb{99.9}          &&95.4      &99.3        &&96.7      &99.4       &&\ul{98.5}    &\ul{99.8}       &&\ul{98.5}    &99.7          &&\tb{98.9}    &\ul{99.8}           \\
  \texttt{Pcb2}    &97.3      &\ul{99.0}          &&93.6      &97.8        &&93.4      &98.0       &&\tb{99.5}    &\tb{99.3}       &&97.6      &98.0          &&\ul{98.7}    &98.7             \\
  \texttt{Pcb3}    &97.9      &\ul{99.2}          &&88.6      &98.3        &&92.0      &98.3       &&\ul{98.9}    &\tb{99.4}       &&\tb{99.1}    &98.8          &&98.6      &99.1             \\
  \texttt{Pcb4}    &99.6      &\ul{98.6}          &&99.4      &97.9        &&99.5      &97.7       &&98.9      &\tb{99.1}       &&\ul{99.7}    &\ul{98.6}        &&\tb{99.8}    &\tb{99.1}           \\
  \texttt{Pipe\_fryum}&\ul{99.8}    &99.1            &&97.4      &99.2        &&98.5      &\tb{99.4}    &&99.7      &\ul{99.3}       &&\tb{99.9}    &99.2          &&99.5      &\ul{99.3}           \\\midrule
  \tb{Average}    &95.1      &\ul{98.8}          &&91.9      &98.6        &&93.2      &98.7       &&\tb{98.1}    &\tb{99.1}       &&\ul{97.8}    &\ul{98.8}        &&\tb{98.1}    &\tb{99.1}           \\\bottomrule
  \end{tabular}
\end{table*}

\noindent\tb{Pixel-level Anomaly Localization.}
\Cref{table:PAL,table:PRO} report the results of pixel-level AUROC and PRO, respectively.
As reported in \Cref{table:PAL}, the original \ti{HETMM} achieves the optimal performance (\tb{99.0\%} $\mathtt{AUC_P}$) like PNI \cite{pni} and RealNet \cite{RealNet}, indicating that the proposed \ti{HETMM} can precisely localize almost all the anomalous regions.
Though the performance of 60-sheet \ti{HETMM} (\tb{98.7\%} $\mathtt{AUC_P}$) is slightly inferior to the top tiers, it surpasses the rest by over \tb{0.2\%} $\mathtt{AUC_P}$.
Moreover, as demonstrated in \Cref{table:PRO}, the original \ti{HETMM} (\tb{96.4\%} \texttt{PRO}) favourably outperforms the second-best method PNI \cite{pni} by over \tb{0.4\%} \texttt{PRO}.
Our 60-sheet \ti{HETMM} achieves the thrid-best performance (\tb{95.8\%} \texttt{PRO}), which is slightly lower than PNI \cite{pni} by \tb{0.2\%} \texttt{PRO}.
According to the introduction of PRO metric in \cref{sec:metrics}, the proposed \ti{HETMM} can localize anomalous regions with much lower false-positive rates than competitors.
In other words, our \ti{HETMM} has a better decision boundary compared to the state-of-the-art methods, achieving better anomaly localization quality.
Additionally, 60-sheet \ti{HETMM} is slightly inferior to the original one, suggesting the proposed \ti{PTS} can tremendously maintain the anomaly localization capability.

\noindent\tb{Performance Curves.}
In \cref{fig:curves}, we compare the proposed \ti{HETMM} with 5 state-of-the-art methods: UniAD \cite{UniAD}, PatchCore \cite{patchcore}, HVQT \cite{HVQT}, RD4AD \cite{RD4AD}, and PNI \cite{pni}, on the MVTec AD dataset in terms of 4 performance curves: ROC, PRO, IoU and PR curves.
The results of UniAD\footnote{https://github.com/zhiyuanyou/UniAD}, PatchCore\footnote{https://github.com/amazon-research/patchcore-inspection}, HVQT\footnote{https://github.com/RuiyingLu/HVQ-Trans}, RD4AD\footnote{https://github.com/hq-deng/RD4AD}, and PNI\footnote{https://github.com/wogur110/PNI\_anomaly\_detection} are obtained by their official implementations.
The proposed \ti{HETMM} consistently outperforms all the other methods in those four curves.
The outperformance of ROC, PRO, and IoU curves indicate that \ti{HETMM} can precisely localize anomalous regions with much lower false positive rates.
In addition, \ti{HETMM} surpasses the other advances in the PR curve, indicating that our results' boundaries and anomalous regions are more precise than the competitors, achieving higher precision scores in all the thresholds.
Notably, our method has better discrimination between anomalies and hard-normal examples than other competitors, yielding fewer false positives and missed-detection rates.

\begin{figure*}[!t]
  \centering
  \includegraphics[width=\textwidth]{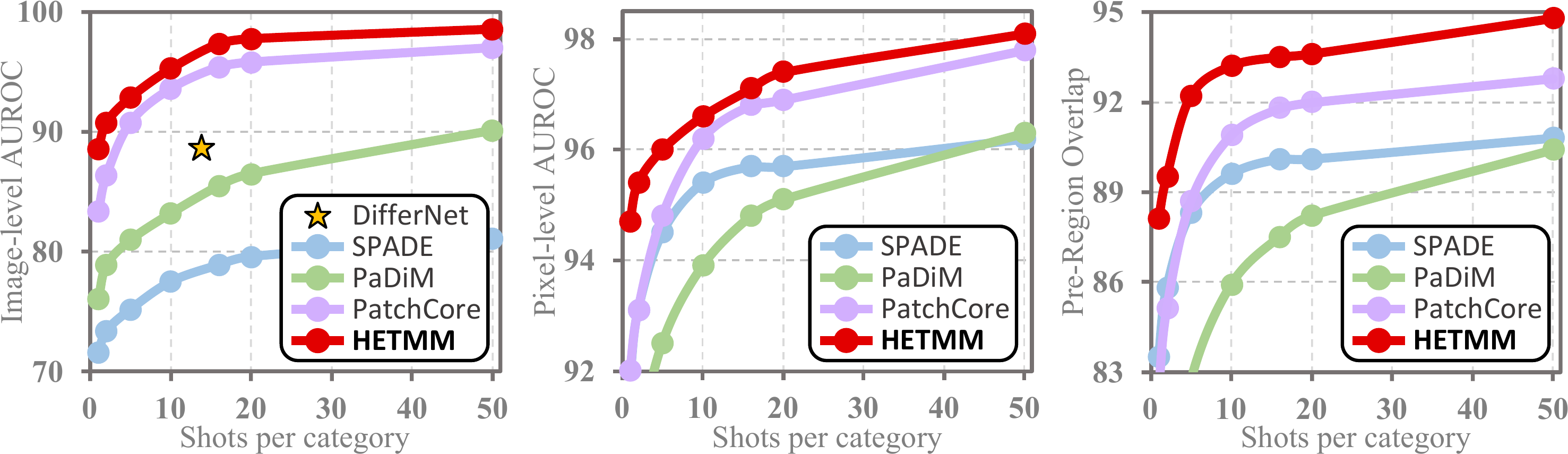}
  \caption{
 The sample-efficiency visualizations for anomaly detection and localization on the MVTec AD \cite{MVTEC} dataset.
 }
  \label{fig:fewshot}
\end{figure*}

\noindent\tb{Visual Comparisons.}
\cref{fig:visual} visualizes the anomaly localization results of the propsoed \ti{HETMM} and other competitors.
It can be observed that the proposed \ti{HETMM} effectively localizes anomalies in various situations, \ti{i.e.,} including anomalies that are too small (rows 1 and 2), anomalies that are too large (rows 3 and 4), multiple anomalies (row 2), and anomalies that reach the borders (row 3).
Compared to the exhibited methods, the segmentation regions of \ti{HETMM} achieve the best visual verisimilitude to the ground truths.
\ti{Note} that \ti{HETMM} can accurately localize the logical anomalies (row 5), while it is a great challenge for the competitors.
In conclusion, \ti{HETMM} achieves better anomaly localization quality, having more precise boundaries and fewer false alarms.

\noindent\tb{Inference Time.}
Though using the original template set achieves the best performance, its inference speed (4.7 FPS) can not satisfy the speed demands in industrial production.
Thanks to the effect of \ti{PTS}, the 60-sheet tiny template set can obtain comparable performance to the original one with over 5x time accelerations (26.1 FPS).

\noindent\tb{Sample Efficiency.}
Sample efficiency is to evaluate the performance of few-shot anomaly detection that detects and localizes anomalies by limited normal samples.
For evaluation, the template set is constructed by varying the number of normal images from 1 to 50, and we randomly construct each template set 10 times to obtain the mean scores.
As shown in \cref{fig:fewshot}, \ti{HETMM} favorably surpasses all the competitors, especially when the template set is constructed with fewer samples.
The outperformance means that \ti{HETMM} shows higher sample-efficiency capability, notably outperforming competitors under the few-shot settings.
Therefore, some incremental issues may be addressed by inserting several novel samples.

\begin{table}[!t]
  \caption{
 Comparisons on the MTD \cite{MTD} dataset in terms of the \tb{AUROC \%} for image-level anomaly detection. 
  \tb{Bold} text indicates the best performance.
 }
  \centering
  \footnotesize
  \begin{tabular}{cccc}
  \toprule        
 PatchCore\cite{patchcore} &CSFlow\cite{csflow} &MSFlow\cite{msflow}  &\multirow{2}{*}{\tb{HETMM}}\\
 (CVPR'22)         &(WACV'22)      &(TNNLS'24)      &\\\midrule
 97.9           &99.3         &99.2         &\tb{99.5} \\\bottomrule
  \end{tabular}
  \label{table:MTD}
\end{table}

\subsection{Evaluation on other datasets} \label{EXP:Others}
We evaluate the performance of \ti{HETMM} on 5 additional anomaly detection benchmarks: The Visual Anomaly (VisA) \cite{VisA}, Magnetic Tile Defects (MTD) \cite{MTD}, Mini ShanghaiTech Campus (mSTC) \cite{mSTC}, MVTec Structural and Logical Anomaly Detection (MVTec SL) \cite{MVTECLOCO} and MVTec Logical Constraints Anomaly Detection (MVTec LOCO) \cite{MVTECLOCO} datasets.
All the images are rescaled into 256$\times$256 resolutions.

\noindent\tb{VisA.}
Following the same one-class classification protocols used in MVTec AD \cite{MVTEC}, we conduct a quantitative comparison against PatchCore \cite{patchcore}, UniAD \cite{UniAD}, HVQT \cite{HVQT}, EffiAD \cite{EffiAD}, and RealNet \cite{RealNet} in terms of image- and pixel-level AUROC on VisA \cite{VisA} dataset.
As reported in \Cref{table:VisA}, the proposed \ti{HETMM} achieves the best average performance for anomaly detection and localization (\tb{98.1\%} $\mathtt{AUC_I}$ and \tb{99.1\%} $\mathtt{AUC_P}$).
\ti{HETMM} also yields consistent preferable performance in all categories, suggesting that \ti{HETMM} has good robustness against various anomalous types.

\noindent\tb{MTD.}
We follow the settings proposed in \cite{DifferNet} to evaluate the anomaly detection performance in terms of the average image-level AUROC against PatchCore \cite{patchcore}, CSFlow \cite{csflow}, and MSFlow \cite{msflow} on the MTD \cite{MTD} dataset in \Cref{table:MTD}.
As shown, the proposed \ti{HETMM} achieves the best performance (\tb{99.5\%} $\mathtt{AUC_I}$), indicating \ti{HETMM} can successfully detect the majority of surface anomalies of magnetic tiles across various illuminative environments.

\noindent\tb{mSTC.}
Following the protocols described in \cref{sec:dataset}, we evaluate the anomaly localization performance in terms of the average pixel-level AUROC against PaDim \cite{Padim}, PatchCore \cite{patchcore}, and MSFlow \cite{msflow} on the mSTC \cite{mSTC} dataset in \Cref{table:mSTC}.
Visually, the proposed \ti{HETMM} achieves the best performance in localizing the region of abnormal events, indicating that \ti{HETMM} can capture the long-range spatiotemporal contexts in videos.
Since mSTC is a non-industrial dataset, the superior performance on the mSTC also suggests \ti{HETMM} not merely works well in industrial situations and has good transferability against various domains.

\begin{table}[!t]
  \caption{
 Comparisons on the mSTC \cite{mSTC} dataset in terms of the \tb{AUROC \%} for pixel-level anomaly localization. 
  \tb{Bold} text indicates the best performance.
 }
  \centering
  \footnotesize
  \begin{tabular}{cccc}
  \toprule
 Padim\cite{Padim}   &PatchCore\cite{patchcore} &MSFlow\cite{msflow} &\multirow{2}{*}{\tb{HETMM}}\\
 (ICPR'21)       &(CVPR'22)         &(TNNLS'24)      &  \\\midrule
 91.2         &91.8           &93.0         &\tb{93.2} \\\bottomrule
  \end{tabular}
  \label{table:mSTC}
\end{table}

\begin{table*}[!t]
  \caption{
 Comparisons on the MVTec SL \cite{MVTECLOCO} dataset in terms of the \tb{AUROC \%} for image-level anomaly detection.
  $\mathcal{S}$ and $\mathcal{L}$ denotes the structural and logical anomalies, respectively.
  \tb{Bold} and \ul{underline} texts indicate the best and second-best performance.
 }
  \renewcommand{\arraystretch}{1.2}
  \centering
  \tiny
  \setlength{\tabcolsep}{1mm}
  \setlength{\belowbottomsep}{3pt}
  \begin{tabular}{c|cccccccccccc}
  \bottomrule 
            &GCAD\cite{MVTECLOCO} &CSFlow\cite{csflow} &PatchCore\cite{patchcore} &RD4AD\cite{RD4AD} &UniAD\cite{UniAD} &PNI\cite{pni} &HVQT\cite{HVQT} &EffiAD\cite{EffiAD}&MSFlow\cite{msflow} &RealNet\cite{RealNet}&\multirow{2}{*}{\tb{HETMM}}\\
            &(IJCV'22)      &(WACV'22)      &(CVPR'22)         &(CVPR'22)     &(NIPS'22)     &(ICCV'23)   &(NIPS'23)    &(WACV'24)     &(TNNLS'24)     &(CVPR'24)      & \\\midrule
  $\mathcal{S}$    &87.1         &98.8        &99.2           &98.7       &96.6       &99.5     &98.1      &99.1        &99.6        &\ul{99.7}      &\tb{99.8}\\
  $\mathcal{L}$    &99.1         &99.0        &92.0           &85.1       &95.8       &\tb{99.9}   &94.5      &98.7        &99.0        &96.1         &\ul{99.8}\\\midrule
  $Avg.$       &93.1         &98.9        &95.6           &91.9       &96.2       &\ul{99.7}   &96.3      &98.9        &99.3        &97.9         &\tb{99.8}\\
  \bottomrule
  \end{tabular}
  \label{table:MVTECSL}
\end{table*}

\noindent\tb{MVTec SL and MVTec LOCO.}
Finally, we follow the protocols reported in \cite{MVTECLOCO} to measure image-level anomaly detection of structural and logical anomalies.
\Cref{table:MVTECSL,table:MVTECLOCO} report the evaluation results on the MVTec SL \cite{MVTECLOCO} and MVTec LOCO \cite{MVTECLOCO} datasets, where $\mathcal{S}$ and $\mathcal{L}$ denote the structural and logical anomalies.
As shown, \ti{HETMM} achieves superior performance to all the competitors on the MVTec SL dataset, indicating that the proposed \ti{HETMM} can precisely identify almost all the structural and logical anomalies on the MVTec AD dataset.
On the MVTec LOCO dataset, \ti{HETMM} achieves the second-best performance, significantly surpassing the other exhibited methods, especially for detecting structural anomalies.
The above performance on the MVTec SL and MVTec LOCO datasets suggests that \ti{HETMM} can capture logical constraints to detect anomalies under the structural and logical settings.
\ti{Note,} \ti{HETMM} is a training-free method, and it can be easily hot-updated by inserting novel samples into the template set.
Compared to the training-based methods: GCAD \cite{MVTECLOCO}, THFR \cite{THFR}, and EffiAD \cite{EffiAD}, \ti{HETMM} can significantly save the training and maintenance costs in practical use.

\subsection{Ablation Study and Sensitivity Analysis}
In this subsection, we first conduct comprehensive ablation experiments to prove the correctness of the designs of \ti{HETMM} and \ti{PTS} with different template matching and template selection strategies, respectively.
Subsequently, we evaluate our method under different hyper-parameters, including pre-trained backbones, patch sizes, hierarchical combinations, and template sizes, to analyze the sensitivity of these hyper-parameters.

\noindent\tb{Different Template Matching Strategies.}
We conduct experiments to prove the correctness of the designs of the proposed \ti{ATMM}.
Specifically, we extensively evaluate the performance of different combinations between template matching and template selection strategies on the MVTec AD \cite{MVTEC} dataset in \Cref{table:RTMM}, where ``\textit{Pixel.}'', ``\textit{Patch.}'', ``$\mathcal{F}$-\tb{ATM}'', and ``$\mathcal{B}$-\tb{ATM}'' denote the pixel-level template matching, patch-level template matching, forward \ti{ATM}, backward \ti{ATM}, respectively.
As shown, pixel-level template matching has inferior robustness and achieves the worst performance across various template selection strategies except applying pixel-level selection strategies on patch-level template matching\footnote{Because each feature selected by pixel-level selection strategies may come from different images, this alters the intrinsic layout of patch-level features (see \cref{fig:overall}).\label{fn:ptm}}.
Though patch-level template matching achieves acceptable performance in an image- and pixel-level AUROC, due to the affection brought by easy-normal examples, its performance is
inferior to forward and backward ATM, especially in \texttt{PRO}, suggesting our analysis in \cref{fig:HAM} and \Cref{sec:HAM} is correct.
The proposed \ti{ATMM} performs better than forward and backward ATM modules across various template selection strategies, indicating that the anomalous signals captured by forward and backward ATM modules complement each other.
Thus, the forward and backward \ti{ATM} modules can be mutually promoted by each other, demonstrating the correctness of our motivation and model design.

\begin{table}[!t]
  \caption{
 Comparisons on the MVTEC LOCO \cite{MVTECLOCO} dataset in terms of the \tb{AUROC \%} for image-level anomaly detection.
  $\mathcal{S}$ and $\mathcal{L}$ denotes the structural and logical anomalies, respectively.
  \tb{Bold} and \ul{underline} texts indicate the best and second-best performance.
 }
  \tiny
  \centering
  \tiny
  \setlength{\tabcolsep}{1mm}
  \begin{tabular}{c|ccccc}
    \toprule
    \multirow{2}{*}{} &GCAD\cite{MVTECLOCO}&PatchCore\cite{patchcore} &THFR\cite{THFR} &EffiAD\cite{EffiAD} &\multirow{2}{*}{\tb{HETMM}} \\
             &(IJCV'22)      &(CVPR'22)         &(ICCV'23)    &(WACV'24)      & \\\midrule
    $\mathcal{S}$   &80.6        &84.8           &86.7         &\tb{94.1}         &\ul{92.9}    \\
    $\mathcal{L}$   &\tb{86.0}      &75.8           &85.2         &\ul{85.8}         &83.2      \\\midrule
    $Avg.$      &83.3        &80.3           &86.0         &\tb{90.0}         &\ul{88.1}    \\\bottomrule
  \end{tabular}
  \label{table:MVTECLOCO}
\end{table}

\begin{table*}[!t]
  \caption{
 Performance of the combinations between various template selection and matching strategies on the MVTec AD \cite{MVTEC} dataset in terms of $\mathtt{AUC_I}$ / $\mathtt{AUC_P}$ / \texttt{PRO}. 
    \texttt{ALL} indicates the original template set.
    \texttt{IR} and \texttt{PR} denote the image- and pixel-level random selection, while \texttt{IC} and \texttt{PC} denote the image- and pixel-level cluster centres, respectively.
 The subscripts of $K$ and $O$ indicate $K$-Means \cite{kmeans} and OPTICS \cite{OPTICS}, respectively.
 The size of each tiny set is set to $60$ sheets as default.
    \tb{Bold} text indicates the best performance.
 }
  \label{table:RTMM}
  \setlength{\tabcolsep}{3mm}
  \centering
  \footnotesize
  \begin{tabular}{lccccc}
  \toprule
  &\textit{Pixel.}    &\textit{Patch.}     &$\mathcal{F}$-\tb{ATM}      &$\mathcal{B}$-\tb{ATM}   &\tb{ATMM}            \\\midrule
  \texttt{ALL}    &96.6 / 97.8 / 94.0   &98.5 / 98.0 / 94.9    &99.2 / 98.4 / 96.0        &99.0 / 98.3 / 95.4     &\tb{99.8} / \tb{99.0} / \tb{96.4}\\\midrule
  \texttt{IR}     &93.3 / 96.4 / 93.0   &95.9 / 97.0 / 94.4    &97.2 / 97.2 / 94.4        &96.4 / 97.0 / 93.8     &98.3 / 97.9 / 93.9        \\
  \texttt{PR}     &93.6 / 96.5 / 93.3   &91.8 / 95.8 / 93.0    &97.6 / 97.4 / 94.6        &96.1 / 96.9 / 94.1     &98.4 / 97.9 / 94.1        \\
  \texttt{IC}$_K$   &93.5 / 96.0 / 93.2   &95.2 / 97.2 / 94.5    &96.8 / 96.9 / 93.3        &95.5 / 96.2 / 93.2     &97.1 / 97.6 / 94.1        \\
  \texttt{IC}$_O$   &92.8 / 95.8 / 93.2   &95.3 / 97.1 / 94.5    &96.9 / 96.8 / 93.4        &95.3 / 96.1 / 93.1     &96.6 / 97.6 / 93.8        \\
  \texttt{PC}$_K$   &95.4 / 96.5 / 93.6   &92.4 / 95.5 / 93.4    &98.3 / 97.7 / 95.0        &97.0 / 97.3 / 94.5     &98.7 / 98.0 / 94.7        \\
  \texttt{PC}$_O$   &95.1 / 96.4 / 93.5   &92.6 / 95.6 / 93.5    &98.0 / 97.6 / 95.0        &97.1 / 97.2 / 94.4     &98.9 / 98.1 / 94.5        \\\rowcolor{MyGray}
  \texttt{PTS}$_K$  &96.3 / 97.5 / 93.9   &93.0 / 95.8 / 93.6    &98.7 / 98.2 / 95.4        &98.6 / 97.7 / 95.0     &99.3 / \tb{98.7} / 95.7      \\\rowcolor{MyGray}
  \texttt{PTS}$_O$  &96.5 / 97.6 / 93.9   &92.8 / 95.7 / 93.5    &98.9 / 98.3 / 95.5        &98.8 / 98.0 / 95.3     &\tb{99.5} / \tb{98.7} / \tb{95.8}\\\bottomrule
  \end{tabular}
\end{table*}

As reported in \cite{MVTECLOCO}, the MVTec AD dataset contains few logical anomalies.
Thus, to evaluate the advantages of \ti{ATMM} in detecting logical anomalies, we further compare \ti{ATMM} with the forward and backward \ti{ATM} on the MVTec LOCO \cite{MVTECLOCO} dataset.
As reported in \Cref{table:LOCO}, forward \ti{ATM} achieves superior performance in detecting structural anomalies but is inferior to backward \ti{ATM} in detecting logical anomalies.
Structural and logical anomalies can be well-detected by the forward and backward \ti{ATM}, respectively.
Hence, by combining these bi-directional matching, \ti{ATMM} achieves the optimal average performance.

\noindent\tb{Different Template Selection Strategies.}
To evaluate the effect of the proposed \ti{PTS}, we conduct quantitative comparisons between the proposed \ti{PTS} and other template selection strategies across various template matching strategies in \Cref{table:RTMM}, where ``$\texttt{ALL}$'' denotes the original template set, and the rest ones are the tiny set with 60 sheets, including random selection: image-level random selection ``\texttt{IR}'', and pixel-level random selection ``\texttt{PR}''; and cluster centres: image-level cluster centres ``$\mathtt{IC}$'', and pixel-level cluster centres ``$\mathtt{PC}$''. 
The subscripts $K$, and $O$ represent $K$-Means \cite{kmeans} and OPTICS \cite{OPTICS}, respectively.
Visually, the performance of \texttt{IC} and \texttt{IR} is unsatisfactory because image-level features contain many redundant information.
The tiny sets built by \texttt{PR}, $\mathtt{PC}_K$ and $\mathtt{PC}_O$ yield better performance than the image-level template selections across various template matching strategies except for the patch-level template matching\footref{fn:ptm}, suggesting that the redundant degrees can be significantly reduced by the selection in pixel levels.
However, as discussed in \Cref{sec:pts}, the tiny sets built by cluster centres struggle to preserve the original decision boundary as they overlook the coverage of hard-normal examples.
As a result, $\mathtt{PTS}_K$ and $\mathtt{PTS}_O$ further improve the performance across various template matching strategies, demonstrating the effectiveness of the proposed hard-normal example selection module.

\begin{figure*}[!t]
  \centering
  \includegraphics[width=\textwidth]{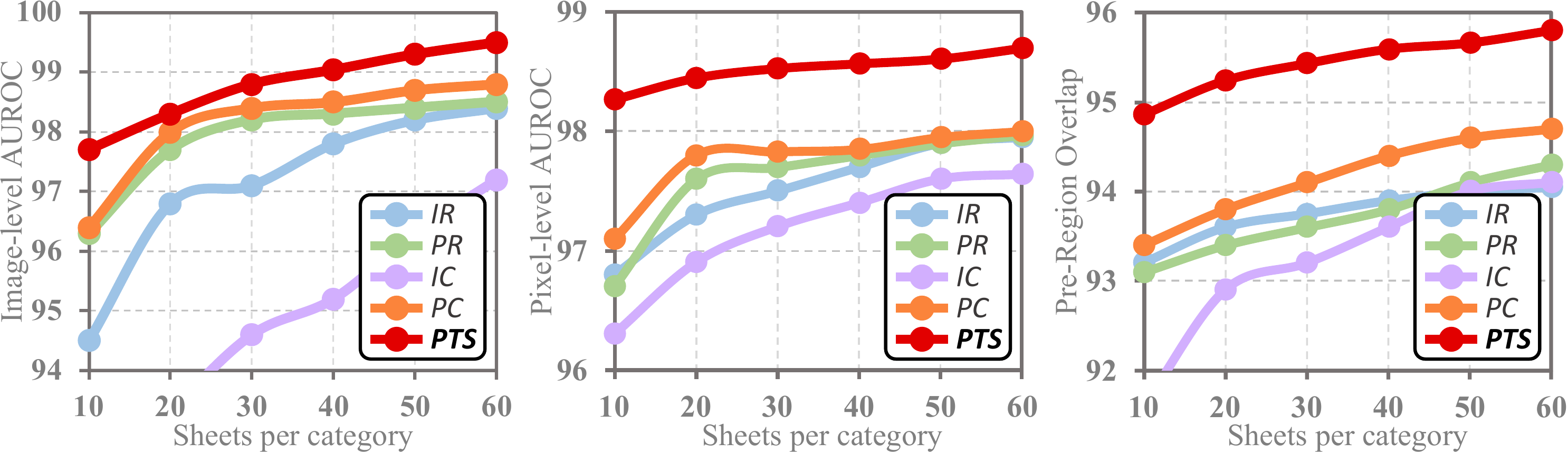}
  \caption{
 The multi-prototype-efficiency visualizations for anomaly detection and localization on the MVTec AD \cite{MVTEC} dataset. $IC$ and $PC$ are generated by collecting the cluster centres of $K$-Means.
 }
  \label{fig:template}
\end{figure*}

We also compare \ti{PTS} and other template selection strategies, including $IR$, $PR$, $IC$, and $PC$.
The cluster method here is K-Means \cite{kmeans} re-implemented by faiss \cite{faiss}.
\cref{fig:template} visualizes the multi-prototype-efficiency for anomaly detection and localization on the MVTec AD \cite{MVTEC} dataset.
Visually, \ti{PTS} favorably outperforms the competitors over various template set sizes, indicating that the proposed \ti{PTS} achieves the best sample-efficiency against the compared template selection strategies, which can better preserve the original decision boundary with the same template size.

The above outperformance demonstrates that the proposed \ti{PTS} achieves better performance across various template matching strategies and template set sizes, proving the validity of our motivation and model design.
Moreover, this also reveals the effectiveness of the hard-normal selection in distribution representation, while it has received little attention from previous literature.

\begin{table}[!t]
  \caption{
 Image-level AUROC comparisons between \ti{ATMM} and its sub-modules: forward \ti{ATM} ($\mathcal{F}$-\tb{ATM}) and backward \ti{ATM} ($\mathcal{B}$-\tb{ATM}), on the MVTec LOCO \cite{MVTECLOCO} dataset. 
    \tb{Bold} and \ul{underline} texts indicate the best and second-best performance.
 }
  \centering
  \footnotesize
  \begin{tabular}{c|ccc}
  \toprule
         &$\mathcal{F}$-\tb{ATM}   &$\mathcal{B}$-\tb{ATM}  &\tb{ATMM} \\\midrule
  $\mathcal{S}$ &\tb{93.4}          &81.8              &\ul{92.9}    \\
  $\mathcal{L}$ &78.8            &\tb{84.0}           &\ul{83.2}    \\\midrule
  $Avg.$    &\ul{86.1}          &82.9              &\tb{88.1} \\\bottomrule
  \end{tabular}
  \label{table:LOCO}
\end{table}

\noindent\tb{Different pre-trained backbones.}
Since the proposed \ti{ATMM} only utilizes the pre-trained parameters of WResNet101 \cite{wresnet} without training and finetuning, its outperformance may be entirely coincidental.
To address this concern, we conduct the quantitative comparison to evaluate the performance of \ti{ATMM} over different pre-trained backbones, including three types of pre-trained backbones in the ResNet family: ResNet \cite{resnet}, ResNeXt \cite{resnext} and WResNet \cite{wresnet}; ``tiny'', ``small'', and ``base'' pre-trained models from the Swin Transformer \cite{swin} and its advanced version \cite{swinv2}; ViT-B/16 and ViT-L/16 based on the ViT architectures \cite{ViT}; and the ``tiny'', ``small'', ``base'', and ``large'' pre-trained models of ConvNeXt-based architectures \cite{convnext}.
\textit{Note}, for ResNet-, Swin-, and ConvNeXt-based backbones, we select the hierarchical features to build the query and template set.
For ViT-based backbones, we resize the images into $224\times224$ and select the features at the final layer to build them.
As shown in \Cref{table:backbones}, \ti{ATMM} performs well on ResNet- and Swin-based architectures, which can achieve comparable performance to the state-of-the-art methods exhibited in \Cref{table:IAD,table:PAL,table:PRO} using a light-weight model such as ResNet18 \cite{resnet} and Swin-T \cite{swin}.
When using ViT- and ConvNeXt-based architectures, though \ti{ATMM} can achieve acceptable performance in image- and pixel-level AUROC, it struggles to localize the anomalies at a low false-positive rate, resulting in unsatisfactory \texttt{PRO} scores.
Since the performance of WResNet101 is optimal for all the metrics, we employ WResNet101 for anomaly detection as default.

\begin{table}[!t]
  \setlength{\tabcolsep}{1mm}
  \caption{
 Results of different pre-trained backbones on the MVTec AD \cite{MVTEC} dataset in terms of $\mathtt{AUC_I}$ / $\mathtt{AUC_P}$ / \texttt{PRO}.
    \tb{Bold} text indicates the best performance.
 }
  \centering
  \tiny
  \label{table:backbones}
  \begin{tabular}{lccc}
  \toprule
  \footnotesize\texttt{Backbones}           &\footnotesize Texture    &\footnotesize Object     &\footnotesize Average \\\midrule
  \multicolumn{4}{l}{\footnotesize\ti{ResNet-based Architectures}}\\
  \footnotesize\texttt{ResNet18}\cite{resnet}     &99.8/97.7/95.1        &99.0/98.0/95.1        &99.3/97.9/95.1   \\
  \footnotesize\texttt{ResNet50}\cite{resnet}     &\tb{99.9}/98.2/95.5     &99.5/98.2/95.5        &99.6/98.2/95.5   \\
  \footnotesize\texttt{ResNet101}\cite{resnet}    &99.8/98.2/95.8        &99.3/98.5/96.2        &99.5/98.4/96.1   \\
  \footnotesize\texttt{ResNeXt50}\cite{resnext}    &\tb{99.9}/98.1/95.6     &99.4/97.8/95.0        &99.6/97.9/95.2   \\
  \footnotesize\texttt{ResNeXt101}\cite{resnext}   &99.8/98.2/96.0        &99.5/98.3/96.0        &99.6/98.3/96.0   \\
  \footnotesize\texttt{WResNet50}\cite{wresnet}    &\tb{99.9}/98.3/95.7     &99.3/98.0/95.5        &99.5/98.1/95.6   \\
  \footnotesize\texttt{WResNet101}\cite{wresnet}   &\tb{99.9}/\tb{98.7}/\tb{96.2}&\tb{99.7}/\tb{99.1}/\tb{96.5}&\tb{99.8}/\tb{99.0}/\tb{96.4}\\\midrule
  \multicolumn{4}{l}{\footnotesize\ti{ViT-based Architectures}}\\
  \footnotesize\texttt{ViT-B/16}\cite{ViT}    &98.5/95.9/89.2        &96.2/98.0/88.5        &97.0/97.3/88.7   \\
  \footnotesize\texttt{ViT-L/16}\cite{ViT}    &98.8/97.2/91.4        &96.0/96.3/87.7        &96.9/96.6/89.0   \\\midrule
  \multicolumn{4}{l}{\footnotesize\ti{Swin-based Architectures}}           \\
  \footnotesize\texttt{Swin-T}\cite{swin}    &99.4/98.0/95.6        &98.3/98.8/94.0        &98.7/98.6/94.6   \\
  \footnotesize\texttt{Swin-S}\cite{swin}    &99.8/98.2/96.2        &97.9/98.7/92.9        &98.5/98.5/94.0   \\
  \footnotesize\texttt{Swin-B}\cite{swin}    &99.5/98.1/95.6        &97.9/98.7/92.8        &98.4/98.5/93.7   \\
  \footnotesize\texttt{SwinV2-T}\cite{swinv2}  &99.4/98.4/96.0        &98.5/96.4/89.8        &98.8/97.0/91.9   \\
  \footnotesize\texttt{SwinV2-S}\cite{swinv2}  &99.2/98.3/95.7        &98.2/97.8/91.5        &98.5/97.9/92.9   \\
  \footnotesize\texttt{SwinV2-B}\cite{swinv2}  &99.5/98.2/95.9        &98.3/96.5/90.0        &98.7/97.1/92.0   \\\midrule
  \multicolumn{4}{l}{\footnotesize\ti{ConvNeXt-based Architectures}}         \\
  \footnotesize\texttt{ConvNeXt-T}\cite{convnext}&94.7/96.5/92.0        &95.7/97.3/90.6        &95.4/97.0/91.1   \\
  \footnotesize\texttt{ConvNeXt-S}\cite{convnext}&98.5/97.6/94.2        &96.4/97.0/90.3        &97.1/97.2/91.6   \\
  \footnotesize\texttt{ConvNeXt-B}\cite{convnext}&97.1/97.2/93.6        &96.6/98.0/91.5        &96.8/97.7/92.2   \\
  \footnotesize\texttt{ConvNeXt-L}\cite{convnext}&98.3/97.3/94.0        &96.9/97.4/90.9        &97.4/97.4/91.9   \\\bottomrule
  \end{tabular}
\end{table}

\begin{figure*}[!t]
  \centering
  \includegraphics[width=\textwidth]{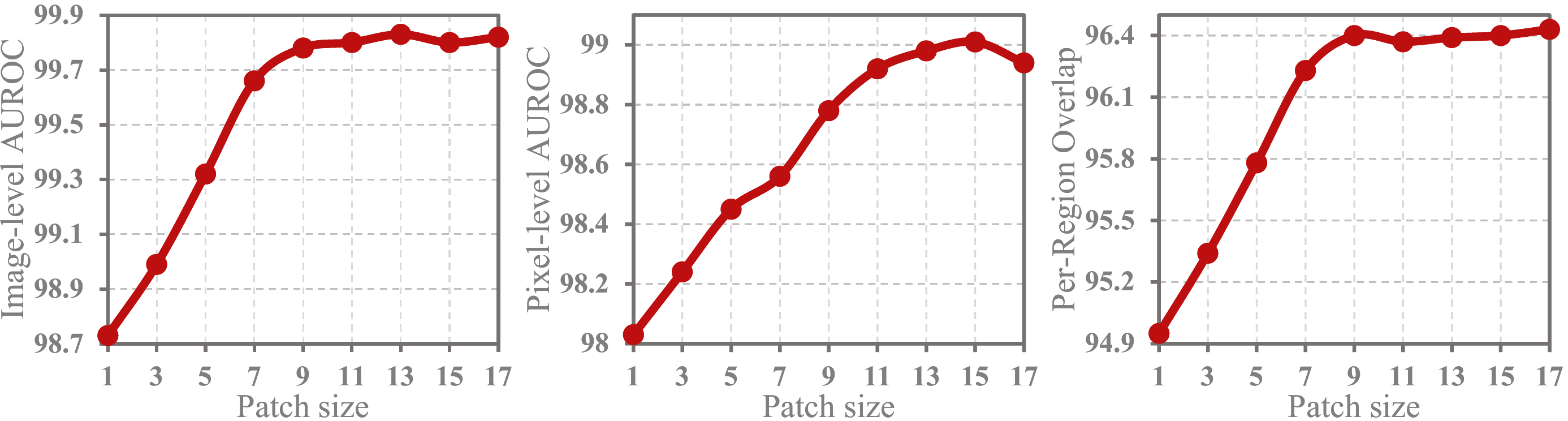}
  \caption{
 The performance over different patch sizes on the MVTec AD \cite{MVTEC} dataset.
 }
  \label{fig:ps}
\end{figure*}

\noindent\tb{Different hierarchical combinations.}
We explore the effectiveness of hierarchical contexts by evaluating the anomaly detection performance over different combinations of the hierarchies in \cref{eq:final}.
The results in \Cref{table:hierarchies} highlight an optimum of the hierarchy combinations with the fixed patch sizes (9$\times$9, 7$\times$7 and 5$\times$5 correspond to hierarchical layers 1, 2, and 3).
As shown, features from hierarchies 1+2 can surpass most existing advances but benefit from global contexts captured by more profound hierarchical combinations 1+2+3.

\noindent\tb{Different patch sizes.}
We investigate the importance of spatial search areas in \cref{sec:HAM} by evaluating changes in anomaly detection performance over different patch sizes in \cref{eq:PT}.
To simplify the evaluations, we only change the patch sizes of hierarchical layer 1.
Results in \cref{fig:ps} show that the performance slightly improves with the increased patch sizes.
As reported in PatchCore \cite{patchcore}, due to the distraction brought by easy-normal examples, its performance first increases and then decreases with the elevated patch sizes.
Hence, the consistent outperformance with the increased patch sizes indicates that such an issue can be addressed by the proposed \ti{ATMM}.

\begin{table}[!t]
  \caption{
 Results on the MVTec AD \cite{MVTEC} dataset over different combinations of hierarchies (from \texttt{layer 1} to \texttt{3}). \tb{Bold} text indicates the best performance.
 }
  \centering
  \label{table:hierarchies}
  \begin{tabular}{ccc|ccc}
  \toprule
  \texttt{layer 1}& \texttt{layer 2}&\texttt{layer 3} &$\mathtt{AUC_I}$ &$\mathtt{AUC_P}$ &\texttt{PRO} \\\midrule
  $\checkmark$  &         &         &97.3       &95.1       &92.7     \\
          &$\checkmark$   &         &98.9       &96.9       &94.5     \\
          &         &$\checkmark$   &96.4       &95.9       &88.4     \\
  $\checkmark$  &$\checkmark$   &         &99.0       &97.1       &95.0     \\
          &$\checkmark$   &$\checkmark$   &98.8       &98.0       &94.3     \\
  $\checkmark$  &         &$\checkmark$   &98.7       &97.9       &94.6     \\
  $\checkmark$  &$\checkmark$   &$\checkmark$   &\tb{99.3}     &\tb{98.2}     &\tb{95.4}  \\\bottomrule
  \end{tabular}
\end{table}

\noindent\tb{Different template sizes.}
\Cref{table:Templates} reports the relationship between the performance (PRO, image- and pixel-level AUROC), template set size, and inference time of our method on the MVTec AD dataset.
Compared to the original template set, the 60-sheet tiny template set compressed by the proposed \ti{PTS} achieves comparable performance with a breakneck speed (26.1 FPS).
Besides, by increasing the template set size from 10 to 60, the inference time has a minor increase, and the performance rises to different extents.
Image-level AUROC, an anomaly detection metric, significantly increases by 2.4\%, while the anomaly localization metrics: pixel-level AUROC and PRO, slightly rise by 0.5\% and 1.0\%, respectively.
Combined with the results reported in \Cref{table:IAD}, the detection of some categories, like the screws, is vulnerable to failure brought by the incompleteness of the template set.

\subsection{Summary}
For \ti{ATMM}, as depicted in \cref{fig:HAM}, it can precisely discover the corresponding prototypes from the template set without the distraction brought by easy-normal examples, achieving much lower false positives than others on MVTec AD \cite{MVTEC} dataset (see \Cref{table:PRO} and \cref{fig:curves}).
The results in \Cref{table:RTMM} also confirm the correctness of our motivation.
To detect the logical anomalies reported in \cite{MVTECLOCO}, \ti{ATMM} combines bi-directional \ti{ATM} to capture structural differences and logical constraints.
As shown in \Cref{table:LOCO,table:MVTECSL}, the outperformance of \ti{ATMM} benefits from the design of bi-directional matching.

\begin{table}[!t]
  \setlength{\tabcolsep}{3mm}
  \caption{
 Relationship between the performance ($\mathtt{AUC_I}$, $\mathtt{AUC_P}$ and \texttt{PRO}), template size (from \texttt{10} to \texttt{60}), and inference speed (\texttt{FPS}) of the proposed \ti{HETMM} on the MVTec AD \cite{MVTEC} dataset.
 }
  \centering
  \scriptsize
  \label{table:Templates}
  \begin{tabular}{l|cccc}
    \toprule
    \texttt{Sheets} & $\mathtt{AUC_I}$ & $\mathtt{AUC_P}$ & \texttt{PRO} & \texttt{FPS} \\\midrule
    \texttt{10}   &97.1       &98.2       &94.8     &31.7\\
    \texttt{20}   &98.3       &98.3       &95.2     &31.2\\
    \texttt{30}   &98.8       &98.4       &95.4     &30.5\\
    \texttt{40}   &99.1       &98.5       &95.6     &29.5\\
    \texttt{50}   &99.3       &98.5       &95.7     &28.0\\
    \texttt{60}   &99.5       &98.7       &95.8     &26.1\\
    \texttt{ALL}  &99.8       &99.0       &96.4     &4.7\\\bottomrule
  \end{tabular}
\end{table}

For \ti{PTS}, unlike existing methods that only capture the cluster centres to represent the original template set, \ti{PTS} selects cluster centres and hard-normal prototypes to represent easy- and hard-normal distribution, respectively.
\cref{fig:PTD} shows \ti{PTS} achieves better coverage than cluster centres, while the latter only covers the easy-normal distribution.
As shown in \Cref{table:RTMM} and \cref{fig:template}, \ti{PTS} yields consistently preferable performance than other template selection strategies under different template matching strategies and the increasing number of template sheets, respectively.
With the effect of the above-proposed techniques, the proposed \ti{HETMM} framework favorably surpasses existing advances for anomaly detection and localization, especially with much lower false-positive rates than the competitors.

Besides, the sensitivity analysis results demonstrate that the proposed \ti{HETMM} framework has good robustness against different hyper-parameter settings.
Therefore, \ti{HETMM} can achieve superior performance in various situations and can be simply adapted to meet different speed-accuracy demands in practice.

\section{Conclusion}
In this paper, we explore that existing methods are prone to erroneously identifying hard-normal examples as anomalies in industrial anomaly detection, leading to high false-positive rates.
However, this issue is caused by the imbalanced distribution between easy- and hard-normal examples, which has received little attention in previous literature.
To address this issue, we propose a novel yet efficient framework: \ti{HETMM}, which can build a robust prototype-based decision boundary to distinguish hard-normal examples from anomalies.
Specifically, \ti{HETMM} employs the proposed \ti{ATMM} to mitigate the affection brought by the affine transformations and easy-normal examples.
By mutually matching the pixel-level prototypes within the patch-level search spaces between query and template set, \ti{ATMM} can accurately distinguish between hard-normal examples and anomalies, achieving low false-positive and missed-detection rates.
In addition, we also propose \ti{PTS} to compress the original template set for speed-up, enabling the models to meet the speed-accuracy demands in practical production.
\ti{PTS} selects cluster centres and hard-normal examples to preserve the original decision boundary, allowing this tiny set to achieve comparable performance to the original one.
Extensive experiments on six real-world datasets demonstrate that \ti{HETMM} outperforms state-of-the-art methods, while using a 60-sheet tiny set achieves competitive performance and real-time inference speed (around 26.1 FPS) on a single Quadro 8000 RTX GPU.
Since \ti{HETMM} is a training-free method and can be hot-updated by inserting novel samples, it can not only significantly reduce training and maintenance costs but also promptly address some incremental learning issues in industrial manufacturing.

\section*{Acknowledgement}
This project is supported by the Natural Science Foundation of China (No. 62072482).

\bibliographystyle{sn-basic}
\bibliography{sn-bibliography}


\end{document}